\documentclass[11pt,letterpaper]{article}
\usepackage{arxiv}
\usepackage[colorinlistoftodos]{todonotes}
\usepackage{soul}
\usepackage{amsmath}

\usepackage[algoruled,linesnumbered]{algorithm2e}
\usepackage{xcolor}
\usepackage{tcolorbox}
\usepackage{amssymb}
\usepackage{multirow}
\usepackage{subcaption}
\usepackage{pdfpages}
\usepackage{booktabs}
\usepackage[utf8]{inputenc} % allow utf-8 input
\usepackage[T1]{fontenc}    % use 8-bit T1 fonts
\usepackage{hyperref}       % hyperlinks
\usepackage{url}  
\usepackage{amsfonts}       % blackboard math symbols
\usepackage{nicefrac}       % compact symbols for 1/2, etc.
\usepackage{microtype}      % microtypography
\usepackage{graphicx}
\graphicspath{{.}{../}}
\usepackage{multicol, latexsym}
\newcommand{\reals}{\mathbb R}
\usepackage{blindtext}
\usepackage{caption}
\usepackage{tabu}

\usepackage{xspace}

\title{Precarity: Modeling the long term effects of compounded decisions on individual instability}
\author{%
  Pegah Nokhiz\\
  School of Computing\\
  University of Utah\\
  \texttt{pnokhiz@cs.utah.edu} \\
   \And
  Aravinda Kanchana Ruwanpathirana \\
  School of Computing\\
  University of Utah\\
  \texttt{kanchana@cs.utah.edu} \\
  
  \And
  Neal Patwari \\
  McKelvey School of Engineering\\
Washington University in St. Louis\\
  \texttt{npatwari@wustl.edu} \\
  
   \AND
  Suresh Venkatasubramanian \\
  School of Computing\\
  University of Utah\\
  \texttt{suresh@cs.utah.edu} \\
}

\begin{document}
% \nipsfinalcopy is no longer used
\date{}
\maketitle
% \vspace{0.5cm}
\begin{abstract}

When it comes to studying the impacts of decision making, the research has been largely focused on examining the fairness of the decisions, the long-term effects of the decision pipelines, and utility-based perspectives considering both the decision-maker and the individuals. However, there has hardly been any focus on \emph{precarity} which is the term that encapsulates the instability in people's lives. That is, a negative outcome can overspread to other decisions and measures of well-being. Studying precarity necessitates a shift in focus -- from the point of view of the decision-maker to the perspective of the decision subject. This centering of the subject is an important direction that unlocks the importance of parting with aggregate measures to examine the long-term effects of decision making. To address this issue, in this paper, we propose a modeling framework that simulates the effects of compounded decision-making on precarity over time. Through our simulations, we are able to show the heterogeneity of precarity by the non-uniform ruinous aftereffects of negative decisions on different income classes of the underlying population and how policy interventions can help mitigate such effects.
\end{abstract}

\section{Introduction}
\label{sec:introduction}

The study of the social impact of automated decision making has focused largely on issues of fairness at the point of decision, evaluating the fairness (with respect to a population) of a sequence or pipeline of decisions, or examining the dynamics of a game between the decision-maker and the decision subject.  
What is missing from this study is an examination of \emph{precarity}: a term coined by Judith Butler to describe an unstable state of existence in which negative decisions can have ripple effects on one's well-being.  Such ripple effects are not captured by changes in income or wealth alone or by one decision alone.  
To study precarity, we must reorient our frame of reference away from the decision-maker and towards the decision subject; away from aggregates of decisions over a population and towards aggregates of decisions (for an individual) over time. 

An individual who lives with higher precarity is more affected and less able to recover by the same negative decision than another with low precarity. Thus including only the direct impact of a single decision or a few decisions is insufficient to judge if that system was fair.
However, precarity is not an attribute of an individual; it is a result of being subject to greater risks and fewer supports, in addition to starting off at a less secure position.  Precarity is impacted by racism, sexism, ableism, heterosexism, and other systems of oppression, and an individual's intersectional identity may put one at greater risk in society, subject to a lower income for the same job, less able to build wealth even at the same income level, and less able to recover from harm.  

Given that automated decision systems and public policy rules operate in a world in which some people's long term well-being is impacted more by the same action, how do we account for the effects of automated decisions and, more generally, proposed public policy rules?  One may advocate for pilot studies, in which the policy or algorithm is deployed on some group.  However, since precarity is a long-term consequence, a pilot study will necessarily take a long time to evaluate its effects.  When a policy is needed for urgent circumstances, such as addressing the impact of a pandemic, there is little opportunity for testing policies in pilots.

In this paper, we propose a modeling framework to simulate the effects of compounded decisions on an individual over time, incorporating a quantification of their precarity. Our framework allows us to explore the effects of different kinds of decision-making processes on individuals' levels of precarity.  In particular, we are able to demonstrate the ill-effects of compounded decision making on the fairness of automated decisions and policies.  

While our model does not capture the full extent of the realities which place some individuals in the precarious position of being more harmed by the same decision compared to someone in a less precarious state, our model does add sufficient complexity to demonstrate how this can happen, and further, a method to quantify the effect.   The message for fairness advocates is that one must look beyond the effect of a single decision on a large number of people, to look at how aggregates of decisions over time impact individuals as a function of their precarity.

\paragraph{Our contributions:}

The main contributions of this paper can be summarized as follows:
\begin{itemize}
\item We introduce the idea of \emph{precarity} to the world of automated decision making, drawing on an extensive literature in sociology and economics. 
\item We build a simulation framework to experiment with and understand the evolution of precarity in a population. This framework incorporates ideas from macroeconomics as well as the framework of bounded rationality to capture the way income \emph{shocks} affect the long-term dynamics of individual wealth. 
\item We present a suite of insights drawn from our simulation platform that validates some of the observations on precarity we see particularly visible in the context of the encountered (especially in the last year) pandemic and illustrates how we can evaluate the effectiveness of proposed policy interventions. 
\end{itemize}

The simulation framework, along with associated scripts, are available at \url{https://github.com/pnokhiz/precarity}.

%%% Local Variables:
%%% mode: latex
%%% TeX-master: "AIES_2021"
%%% End:

\section{Background and related work}
\label{motivation}

\paragraph{Precarity:}
Precarity \cite{anthro} is a multi-faceted concept that very broadly speaks to the instability and \emph{precariousness} of people's lives. It has been interpreted as an economic condition \cite{nancyworth, greigDepeuter}, a sociological condition that speaks the interconnectedness and therefore vulnerability of human lives \cite{butler2006precarious,butler2016frames}, as a descriptor of a political class characterized by irregular or transient employment \cite{standing2014precariat,gill2016century} or as a psychological condition of exclusion and displacement \cite{allison2014precarious}. 

In this work, our focus is on algorithmic decision systems and the effect of compounded decision-making on households and groups. In that regard, we interpret precarity as the instability associated with sequences of negative decisions: specifically, the way in which repeated negative outcomes can increase the likelihood of one falling into poverty. Ritschard et al.\ \cite{ritschard2018index} were the first to attempt to quantify precarity (in the context of the labor market) by looking at transitions between more or less precarious states (for example, a full time versus a part time job). This work observes that negative transitions have the most critical role in increasing precarity. Aneja et al.\ \cite{aneja2019no} study the effect of incarceration on access to credit -- arguing that incarceration reduces the access to credit, which in turn increases recidivism. An important recent work that has greatly influenced our thinking is by Abebe et al.\ \cite{abebe2020subsidy}. In it, they build a theoretical model to capture the effect of \emph{income shocks} on one's chance of going bankrupt and propose efficient allocations of limited stimulus to maximize the expected number of individuals saved from bankruptcy.

\paragraph{Fairness in sequential decision-making:}

Zhang and Liu \cite{zhang2020fairness} present a comprehensive review of work on fairness in sequential decision making broken down by whether the decision process affects input features or not. 

A large body of work considers the case where input features do not change \cite{heidari2018preventing, gupta2019individual, bechavod2019equal, joseph2018meritocratic, hebert2017calibration, valera2018enhancing, bechavod2019equal, li2019combinatorial, chen2019fair, gillen2018online, patil2019achieving, dwork2018fairness}. When considering how a population might evolve in response to decisions, two broad lines of work emerge -- those that consider two decision stages \cite{liu2018delayed, heidari2019long, downstream} and those that consider finite or infinite-horizon decision making \cite{labor_market_lily_hu, hashimoto2018fairness,zhang2019group,mouzannar,disparat-equi,emelianov2019price}.

This latter body of work is more closely related to our study. The broad goal here is to understand how qualifications of different groups evolve in the long run under various fairness interventions and the conditions to achieve social equality. They often focus on the problem of access and ``dropout'' (when decisions lead to withdrawal from the market) as causes of disparity between groups and propose various interventions to address this. 

Another approach to understanding sequential decision making has been to take advantage of simulations on Markov decision processes (MDP). As \cite{fairness-static} argues, long-term fairness dynamics are hard to evaluate, and so we need simulations to assess fairness over time. MDPs can also be formally analyzed for long-term effects on (group and individual fairness) as explored by \cite{jabbari2017fairness,NIPS2016_6355,wen2019fairness,fairness-static}.

Our work also focuses on simulations to reveal insights about the underlying dynamics. Our focus, however, differs from the above works in that we a) model a system heterogeneously where different states capture different levels of precarity and b) focus on the effect of the system on individual \emph{trajectories} rather than only population-level outcomes. While we do explore fairness concerns, we do this in the context of precarity rather than focusing on tradeoffs between utility and fairness. 
\vspace{-2.0mm}

\paragraph{Economic models of consumption and savings:} Macroeconomics emphasize time-related decisions, such as consumption plans. It is often useful to assume that the time horizon is infinite in these settings, which necessitates the use of dynamic optimization. The infinite horizon models are about optimal consumption and savings  at various points in time, given that the production is subject to random noises (e.g., the optimal growth model) \cite{stokey1989recursive, ljungqvist2018recursive}. For analyzing capital income risk, which is essential for understanding the joint distribution of income and wealth \cite{benhabib2015wealth, stachurski2019impossibility}, there are other infinite horizon models like the Income Fluctuations Model (household consumption and savings problem), which is established in a more dynamic manner where state-dependent returns on assets fluctuate over time \cite{ma2020income}.

% We have to  

\section{Quantifying precarity}
\label{sec:quant-prec}
Precarity (as discussed above) is a broad interdisciplinary notion describing the instability of modern life. In this paper, we focus on the \emph{economic} aspects of precarity -- how financial and other shocks create uncertainty around one's financial status. Within the social sciences, it has long been recognized that standard measures of inequality -- like the Gini index and others -- cannot quantify the dynamics of a precarious trajectory. Indeed, precarity has been referred to as a ``slow death'' \cite{johnson2020precarious, puar2012precarity} because of its progressive nature that unfolds for an individual over time. 

Much research \cite{pelletier2020measuring} has therefore gone into characterizing properties of \emph{sequences} that describe the state of an individual over time. Researchers have proposed measures that seek to capture the \emph{number} of distinct states, the number and direction of transitions between states, and even incorporate the significance and meaning of individual states in the sequence. For example, to capture the variability in states in a sequence, the entropy of the frequency distribution of states has been regularly used. To capture effects at different time scales, other researchers have proposed first constructing subsequences of the trajectory (akin to the use of skip n-gram models in text analysis). In this paper, we use one of these measures, proposed by Ritschard et al.\ \cite{ritschard2018index}, that seeks to capture three key aspects of precarity. We assume that an individual's trajectory is described as a sequence of states $\sigma = s_1, s_2, \dots, s_t$ where $s_i \in S$ and $S$ is the set of states. A \emph{quality} function $r : S \to \reals$ indicates the level of financial wherewithal (where a higher quality implies a better condition). Then the measure of precarity for a given sequence $\sigma$ depends on

\begin{itemize}
    \item The quality of the starting state $r(s_1)$
    \item The net decline in state over $\sigma$
    \item The amount of variability in $\sigma$
\end{itemize}

\paragraph{Net decline in state:}

We assume the states in $S$ are sorted from lowest to highest ``quality''. In any sequence $\sigma$, we can classify transitions between states as either negative or positive, depending on which state is higher. Let $q^-(\sigma)$ be the proportion of transitions that are negative, and $q^+(\sigma)$ be the proportion that are positive, and set $q(\sigma) = q^-(\sigma) - q^+(\sigma)$. The quantity $q(\sigma)$ represents the net magnitude of negative transitions and ranges between $-1$ (for purely positive transitions) and $1$ for purely negative transitions. We note in passing that the transitions can be weighted: in that case, the proportions are appropriately calculated in a weighted manner. We weigh them by the hops (distance) a state has to the least precarious state in a sequence.

\paragraph{The variability in the sequence:}

There are two factors used to define variability in $\sigma$. The first is the number of states visited, or more generally the distribution of the states entered during the sequence. This can easily be captured by computing the entropy $h(\sigma)$ of the (normalized) frequency distribution of states. This in turn, must be normalized by the maximum entropy possible, which is merely $\log |S|$.
This does not however capture the transitions between states. For example, consider the sequences $\sigma = (1,1, 1, 1, 0, 0, 0, 0)$ and $\sigma' = (1,0,1,0,1,0,1,0)$. Clearly $h(\sigma) = h(\sigma')$ but $\sigma'$ reflects a more erratic state of existence. To account for this, Ritschard et al.\ \cite{ritschard2018index} add in a term $t(\sigma)$ that merely counts the number of transitions to different states (normalized by $|\sigma|-1$). Note that $t(\sigma)/(|\sigma|-1) = 1/7$ but $t(\sigma')/(|\sigma|-1) = 1$. These two terms are combined using their geometric mean:

\[ c(\sigma) = \sqrt{\frac{h(\sigma)}{\log |S|} \frac{t(\sigma)}{|\sigma|-1}} \]

\paragraph{The precarity index:}

The overall precarity index of a sequence $p(\sigma)$ is a function of the initial quality $r(s_1)$, the net decline in state $q(\sigma)$ and the amount of variability $c(\sigma)$. In this paper, we use \cite{ritschard2018index}'s formulation of the index: whether other functional forms might provide different sensitivity is a matter we defer to further research. The precarity index can then be defined as: 
\[ p(\sigma) = \lambda r(s_1) + (1 - \lambda) c(\sigma) ^\alpha(1 + q(\sigma))^\gamma \]
This can be seen as a convex combination of the starting position and terms involving dynamic components (controlled by $\lambda$). The two dynamic components are weighted by different exponents to reflect different degrees of sensitivity and importance. We set these values as  $\lambda=0.2$, $\alpha=1$, and $\gamma=1.2$, as is done in \cite{ritschard2018index}. Note that we use the term $1+q(\sigma)$ to yield a term between $0$ and $2$: if the trajectory of the sequence is purely positive (thus setting $q(\sigma) = -1$) the precarity is merely a function of the initial state. In our experiments, we test several values of $\alpha$ and $\gamma$, and they do not affect the results as long as they are above zero, since the overall effects on the underlying population will be similar for all data points. 

We can also now elaborate on why measures like the Gini index fail to capture precarity. Precarity is a notion evaluated for an individual over time -- the precarity index is a way to quantify this as a kind of time average. The Gini index instead is a measure of inequality of a population measured at a snapshot in time and acts as a population aggregate measure.

\section{A simulation based methodology for exploring precarity}  
\label{sec:simul-based-meth}
Continuing in the line of works like \cite{fairness-static} and \cite{zheng2020ai}, we use a  simulation framework to explore the dynamics of precarity. In this simulation framework, individual \emph{agents} make choices (and are subject to decisions) within a system, and are described by parameters for income, wealth and health. We use population-level economic data to initialize the system, and allow the agents to make either locally reasonable decisions (in a \emph{bounded rationality}-like framework) or allow them to maximize expected utility within epochs. Using a simulation framework with realistic input parameters and controls allows us to observe the evolution of the system in a way that would be difficult to do formally (like for example, \cite{abebe2020subsidy} is able to do for the more specific problem of income shocks), and allows us to experiment with different kinds of interventions.   

\paragraph{Agents:} The agents are households who interact with simulated environments in an alternating loop. Each agent is specified by their \textbf{income}, \textbf{net worth}, and \textbf{health}. An agent incurs \textbf{expenses} each time period and also earns income. Agents must make decisions about their assets -- whether to consume, pay for expenses, save, or improve their health.

\paragraph{States:} We associate each agent with a set of three states (one for each of income, net worth, and health). Each state indicates which decile of the overall population they are in for that attribute (so there are a total of $10\times 10\times 10 = 1000$ possible states).

\paragraph{Metrics:}
We use sequences of states for each attribute separately to perform precarity computations for each agent as described in Section~\ref{sec:quant-prec} above. We record the precarity value of all households for each income decile. 

\paragraph{Initialization and updates:}

We initialize a population of agents using parameters drawn from published statistics. For initial income, 
we generate an income distribution of 10,000 points using 2019 income data of the US Census Bureau's Annual ASEC survey of the Consumer Price Index (2019 dollar values) as detailed by the IPUMS Consumer Price Survey \cite{flood2020integrated, DQYDJ}. To each household, we assign a net-worth (their financial and non-financial assets minus their liabilities). The net worth is assigned by detailed median percentile net worth data and median net worth by income by percentile data from the Federal Reserve.\footnote{\url{https://www.federalreserve.gov/}} The health index average of the population is extracted from the Census Bureau CPS Annual Social and Economic (March) Supplement 2019 \cite{census}.\footnote{\url{https://www.census.gov/programs-surveys/cps.html}} Note that we consider one health feature for the entire household. While health is of course a personal state, this allows us to combine this data with the household-based data for the other attributes. 
Each household has a set of basic expenditures each month (e.g., for food, housing, transportation, etc.). These expenses are extracted from 2019 mean annual expenditures from the Consumer Expenditure Surveys of the US Bureau Of Labor Statistics.\footnote{\url{https://www.bls.gov/cex/tables.htm}}

\textbf{Updating health information.} Net worth automatically updates as agents spend their income and/or save it. Income updates happen via a decision process that we describe below. What remains is to describe how the health status updates. The relationship between health and income has been observed to be positive and concave \cite{preston1975changing}. Wagstaff et al.\ \cite{doi:10.1146/annurev.publhealth.21.1.543} have proposed modeling this as a second degree polynomial whose first gradient is positive and the second gradient is negative. To define the function, we use the principle of \emph{relative income theory} where ``health depends
on income relative to average incomes of
one or more reference groups" \cite{deaton2003health}. That is, the individual (we consider the household one entity) health equals income relative to a specific group's income reduced by the square of
income relative to the group's income:  
\[
h_i = \Bar{h} + \eta (w_i - w_g) - \sigma (w_i - w_g)^2,
\]
where $h_i$ is the individual (household's) health, $\Bar{h}$ is the mean health index of the whole population (extracted from CPS), $w_i$ is individual (household)'s income, and $w_g$ is the income mean in the group of reference, i.e., the income decile the household is in, in each round of decision making. $\eta$ and $\sigma$ are positive model parameters. We choose parameters that result in a wider range of indices for precarity states. We choose $1$ and $10^{-20}$ for $\eta$ and $\sigma$, respectively.

\subsection{Income shocks}
\label{sec:income-shocks} 
 
The decisions are made for 10 rounds on a monthly income and expenditure monetary value basis. The effects of positive or negative decisions are reflected on income after each round. We deduct (add) a unit based on negative (positive) outcomes if the households do not stay in the same state. We set this unit as 10\% of their income, which has less financial calamity than setting a fixed value (e.g., \$500) for lower incomes since it decreases proportionally with their income. Clearly, a higher value will be more beneficial for the wealthy and more ruinous for middle and lower income households. 

\textbf{Benefit decision policy.}  Public policy can improve household financial stability by providing benefits, and in these simulations, we explore the effect of decision classifiers which make these decisions based on an individual's current state. We introduce a lenient classifier, which accepts $50\%$ of the initial population applying for the service based on their current income. The threshold is a global fixed value for the whole population despite their previous transitions, highlighting the fact that the decision-maker is unaware of the precariousness of the household. We implemented the experiments for a range of classifier thresholds to see the precarity of the population for the most lenient classifier, the most difficult classifier, and all other classifiers in between. We chose the most lenient classifier to consider the most optimistic scenario for assigning positive decisions. The harsher classifier has the smallest impact on precarity levels. We try to make the default simulation specification in the interest of lower income households.

\subsection{Strategies}
\label{sec:strategies}

The final piece of the simulation is specifying how agents behave at each time step. The economics literature typically views agents as rational (discounted) utility maximizers, and an extensive literature has developed around different stochastic models under which to maximize utility. An alternative approach is to take a viewpoint of \emph{bounded rationality}: each agent now makes realistic choices (stochastically) from a collection of options that are locally rational, but cannot perform long-range utility maximization.

We simulate agent behavior under both of these models, which we describe below. 

\subsubsection{Rational agents and Income Fluctuations Problem (IFP)}

In this model \cite{ma2020income, sargent2014quantitative, deaton1989saving, den2010comparison, kuhn2013recursive, rabault2002borrowing, reiter2009solving, schechtman1977some}, an agent finds a consumption-asset path $ \{(c_t, a_t)\} $ where $a_t$ is the assets (net worth) at point $t$, and $c_t$ is the consumption at point $t$, with the goal of maximizing
\begin{equation}
 \mathbb E \left\{ \sum_{t=0}^\infty \beta^t u(c_t) \right\}   
 \label{obj1}
\end{equation}
such that
\begin{equation}
   a_{t+1} = R_{t+1} (a_t - c_t) + Y_{t+1}
\; \text{ and } \;
0 \leq c_t \leq a_t 
\label{cons}
\end{equation}

Where, $\beta \in (0,1) $ is the discount factor, $Y_t $ is non-capital income (i.e., via labor), and $ R_t $ is the interest rate on savings. For simplicity, in this paper we will disregard gains from savings by setting $R_t = 1$. 

The non-capital income $Y_t$ is controlled by an exogenous state process $z = \{Z_t\}$. As we shall see, this is how we can introduce income shocks via decision processes. 

The quantity $u$ is the utility to the household. We use the Constant Relative Risk Aversion (CRRA) \cite{ljungqvist2018recursive,wakker2008explaining} utility
\[ u(c) = \frac{c^{1 - \gamma_c}} {1 - \gamma_c}, \]
which is a commonly used utility function in finance and economics that captures the idea that risk aversion is independent of scale. Here, risk aversion refers to an individual's inclination to prefer low uncertainty (more predictable) but lower pay results over the results with high uncertainty but higher payoffs \cite{o2018modeling}. We pick $\gamma_c = 2$ since the utility function has a  $c^{(1-\gamma_c)}$ term and with a smaller value, $u(c)$ could become imaginary given that we use $u(c - b)$ where $b$ is their monthly basic expenditures. This is to assure that they cover their basic needs in every round (if $c < b$ then there is negative utility).

A \emph{feasible} consumption path $ (a,z) \in \mathsf S $ is equivalent to the consumption path $ \{c_t\} $. However,  $ \{c_t\} $ and its asset path $ \{a_t\} $ must satisfy the following:
\begin{itemize}

    \item  $ (a_0, z_0) = (a, z) $ 
    \item the feasibility constraints in \ref{cons}
    \item being measurable. This means that only before $t$ (and not afterward) the consumption path is a function of random variables. Thus, at time $t$ the consumption cannot be a function of unobserved outcomes.

\end{itemize}
An \emph{optimal consumption path}  $ (a,z) $ is a feasible consumption path that attains the supremum in $\max{(.)}$ for the objective~\ref{obj1}, which can be shown to be:
\begin{equation}
u' (c_t)
= \max \left\{
    \beta R \,  \mathbb{E}_t  u'(c_{t+1})  \,,\;  u'(a_t)
\right\}  
\label{5}
\end{equation}
This can be optimized to find the optimal consumption. Please see Appendix \ref{sec:supp-optimal-IFP} for details.\footnote{Our explanations and implementation for this model is built upon \url{https://python.quantecon.org/ifp_advanced.html}}

\subsubsection{Markov Decision Process (MDP) model}

We now turn to our second approach to modeling agents. Here, each agent will occupy a state of a Markov decision process, with transitions out of each node based on locally reasonable decisions about asset management. Agents can use their savings to pay for their necessities and liabilities, they can sell a tangible asset, opt for the conversion of a health-related tangible asset (such as an insurance plan). They can use their income to increase their savings, invest in health improvement, or build assets through consumption. The decision they make (stochastically) moves them to a new state with modified attributes (income, wealth and health) accordingly. See Appendix \ref{sec:supp-mcmc} for details on the transitions in this system. We note that the transitions are designed based on prior studies of income and precarity \cite{income-precarity} that describe typical behaviors of individuals in different income classes when faced with income and health shocks.

\paragraph{The decision making process:}

The IFP model presents challenges for the income shock process that we described in Section~\ref{sec:income-shocks}. In the macroeconomic literature on income fluctuation (and indeed also in the work by \cite{abebe2020subsidy}), shocks are assumed to present in a stochastic form. Thus, while there is randomness in the shock generation process, it is a predictable kind that can be optimized for (in an expected sense). However, shocks generated by an external decision cannot be optimized for in the same way (and indeed, this is an important element of precarity). In our simulation, we think of the optimization process as happening in epochs \emph{between} decision points. This model captures the idea that long-range planning is constrained by decision points that the individual has no control over.

\section{An empirical inquiry}
\label{sec:assess_prec}

With our simulation framework now in place, we are ready to explore a set of
questions relating to how precarity manifests itself. For each of these
questions, we will run both simulation methodologies described above in
Section~\ref{sec:strategies}. We will run the simulation for a fixed number of
time steps, recording the (cumulative) precarity indices of individuals for each
of the three state variables as their state string gets longer. We will show the
distribution of precarity index values across the population at each time step
in order to illustrate how the distribution evolves over time.

\subsection{Evolution of precarity}
\label{sec:evolution-precarity}

\begin{figure*}[!htbp]
    \vspace{-1.0mm}
    \centering
    \begin{subfigure}[t]{0.3\textwidth}
        \centering
        \includegraphics[width=\columnwidth]{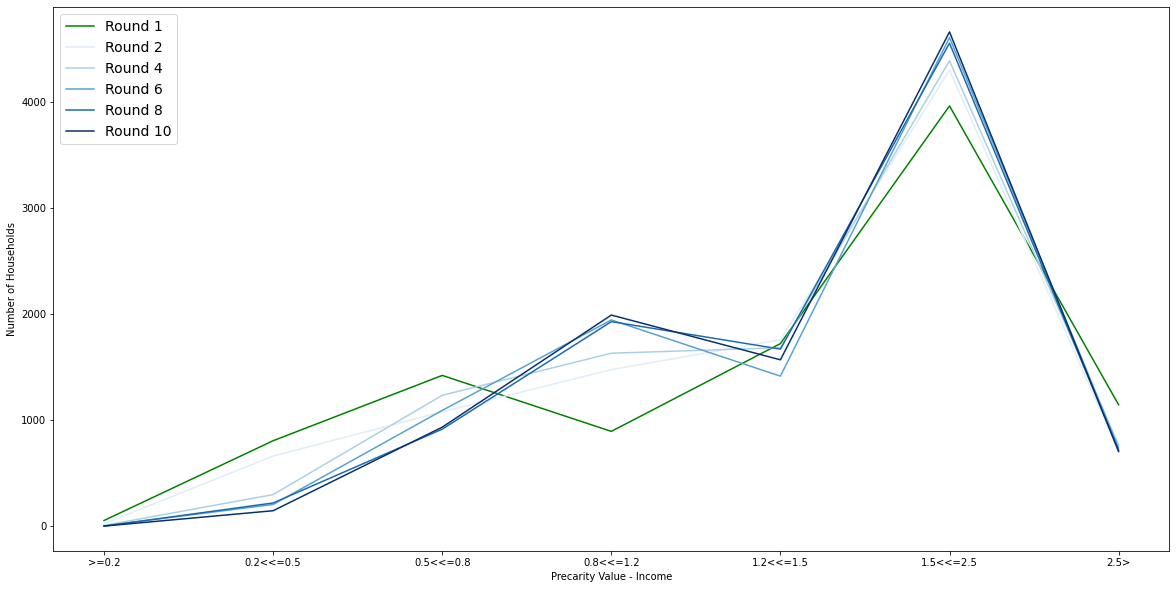}
        \caption{Income precarity}
        \label{fig:ifp-prec_inc_population}
    \end{subfigure} \qquad
    \begin{subfigure}[t]{0.3\textwidth}
        \centering
        \includegraphics[width=\columnwidth]{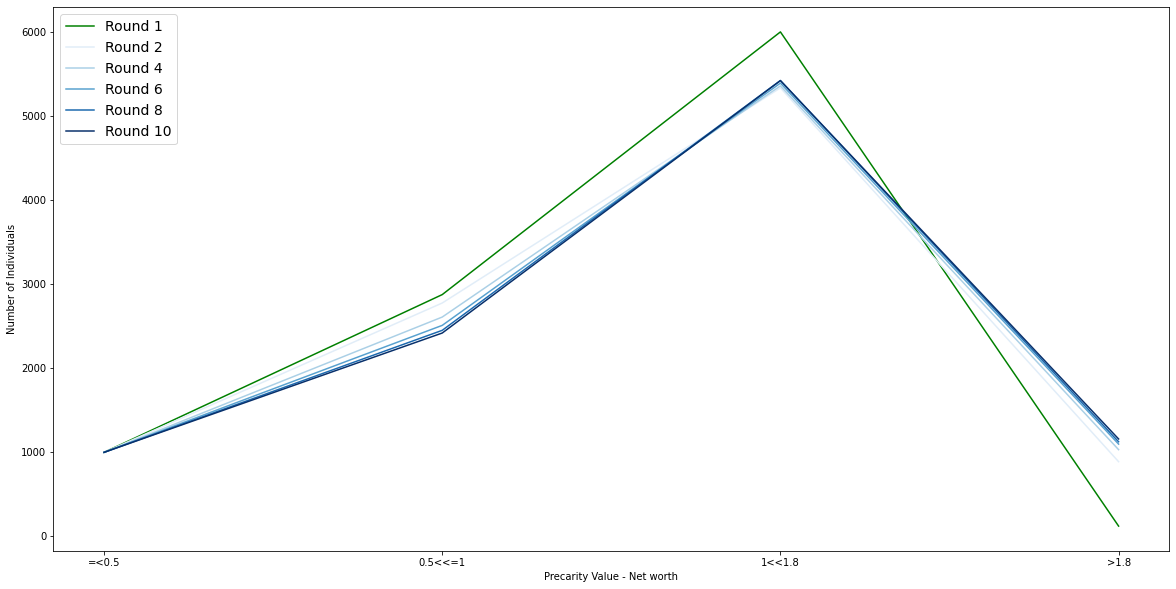}
        \caption{Net worth precarity}
        \label{fig:ifp-prec_savings}
    \end{subfigure}\qquad
    \begin{subfigure}[t]{0.3\textwidth}
        \centering
        \includegraphics[width=\columnwidth]{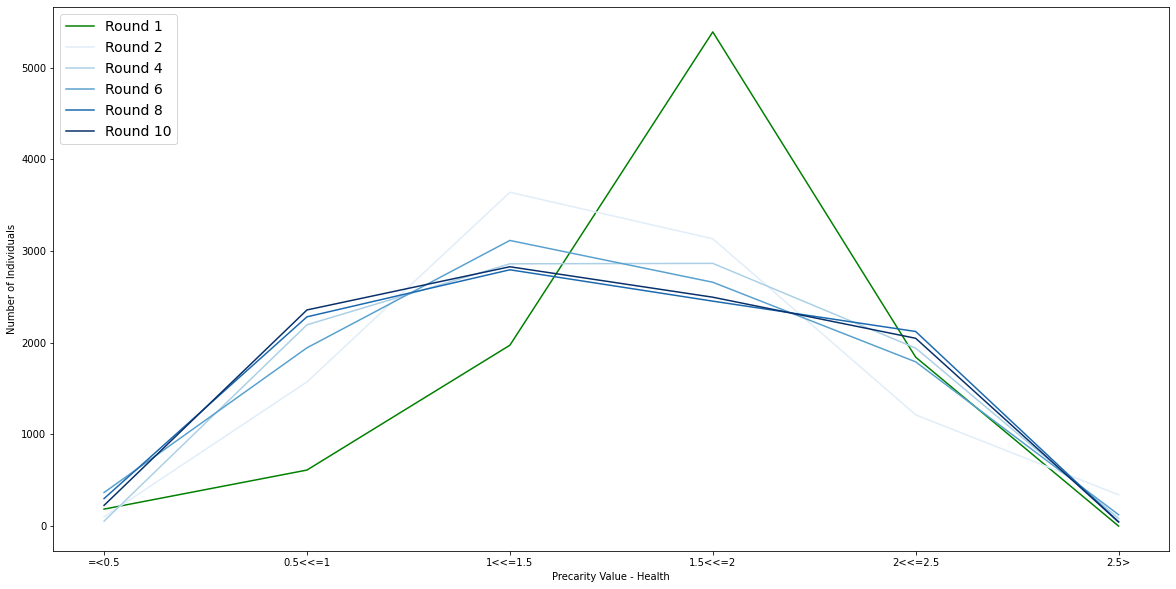}
        \caption{Health index precarity}
        \label{fig:ifp-prec_health}
    \end{subfigure}

    \caption{Assessing household precarity over time - IFP model}
    \label{fig:ifp 10 rounds precarity}        
\end{figure*}

\begin{figure*}[!htbp]
    % \vspace{-1.0mm}
    \centering
    \begin{subfigure}[c]{0.3\textwidth}
        \centering
        \includegraphics[width=\columnwidth]{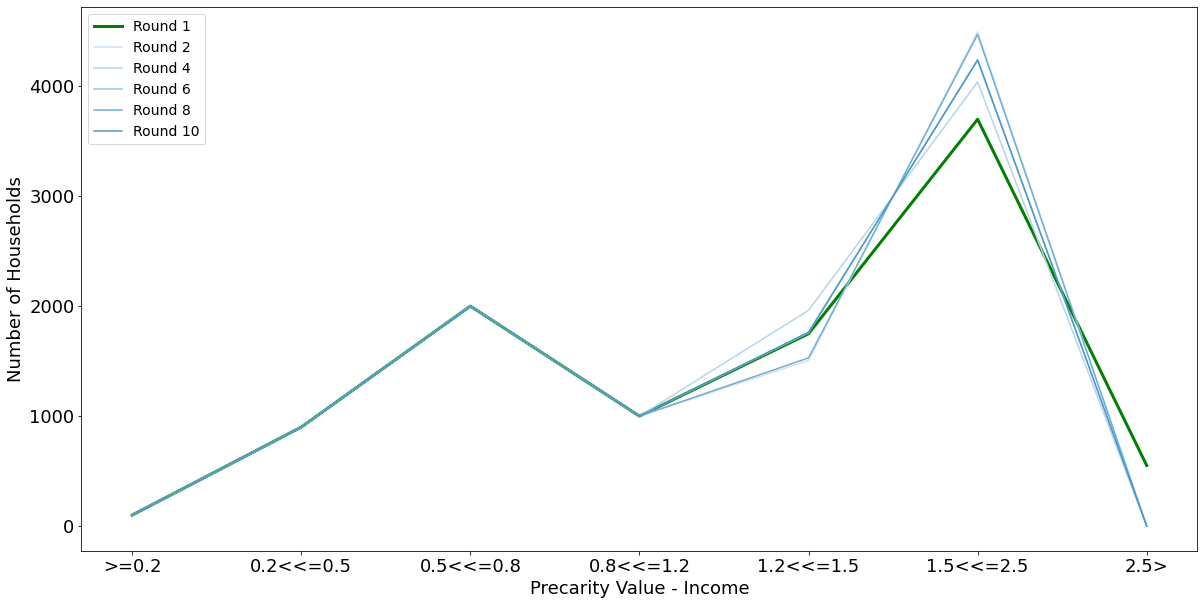}
        \caption{Income precarity}
        \label{fig:prec_inc_population}
    \end{subfigure} \qquad
     \begin{subfigure}[c]{0.3\textwidth}
        \centering
        \includegraphics[width=\columnwidth]{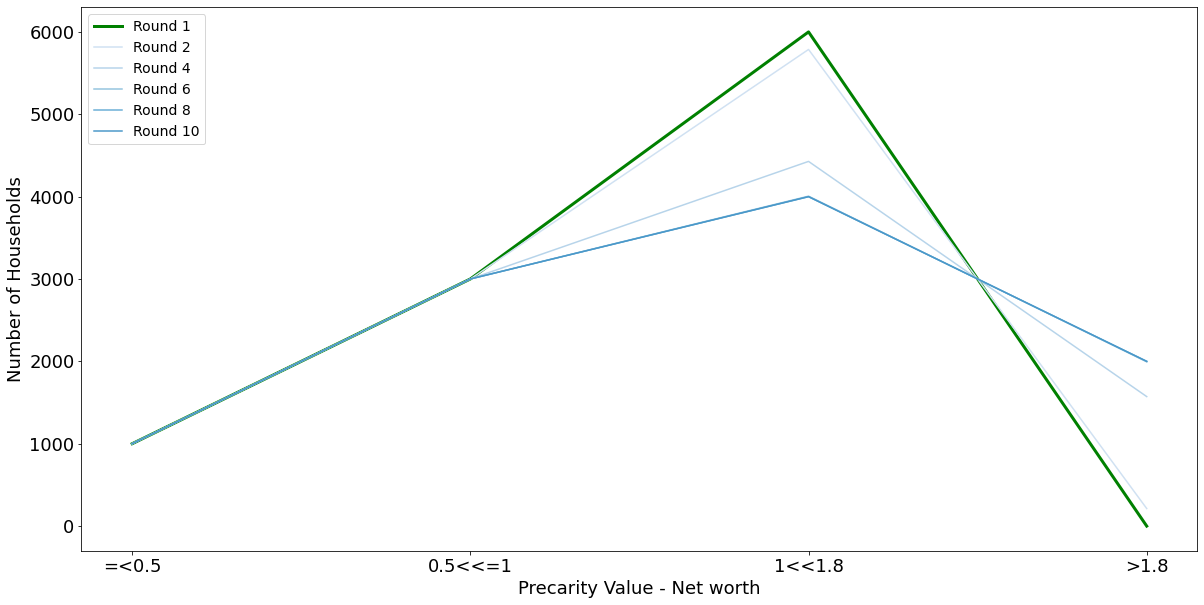}
        \caption{Net worth precarity}
        \label{fig:prec_networth}
    \end{subfigure} \qquad
    \begin{subfigure}[c]{0.3\textwidth}
        \centering
        \includegraphics[width=\columnwidth]{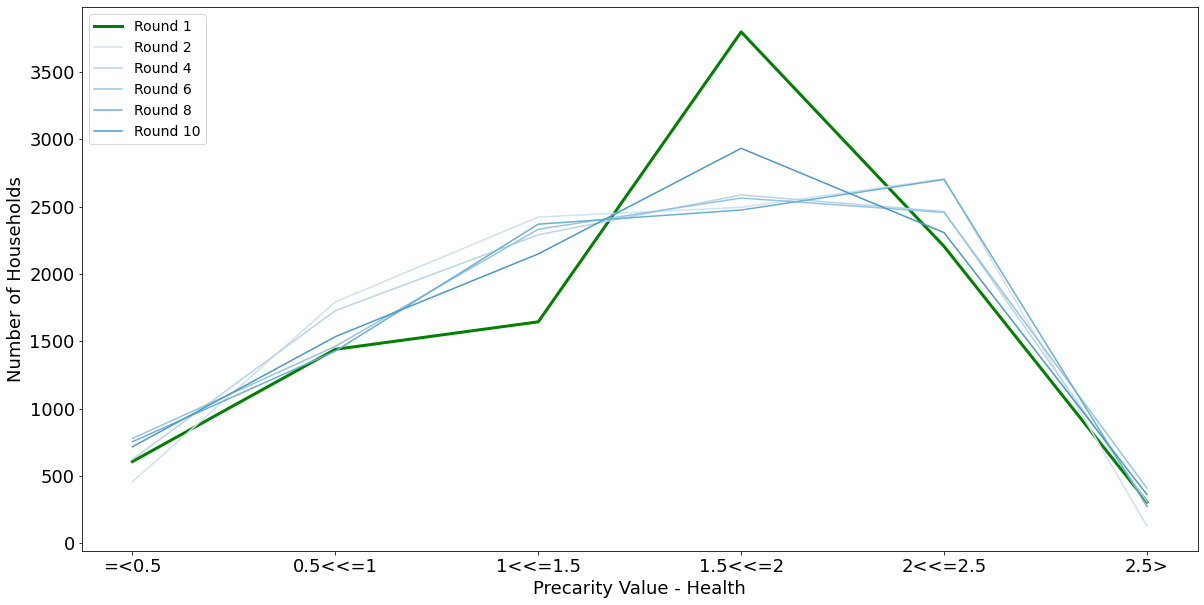}
        \caption{Health index precarity}
        \label{fig:prec_health}
    \end{subfigure}

    \caption{Assessing household precarity over time - MDP Model}
    \label{fig:10 rounds precarity}
     \vspace{-2.0mm}    
\end{figure*}

Our first sequence of experiments acts as a baseline to demonstrate how income
shocks affect the precarity of individuals over time with respect to each of the
three state variables. The results (for each of the variables) for the IFP model
are shown in Figure~\ref{fig:ifp 10 rounds precarity} and the corresponding
results for the MDP model are shown in Figure~\ref{fig:10 rounds precarity}. 

\textbf{Analysis.} 
In both models we observe that as the system progresses the precarity
distribution for  net worth shifts rightward (i.e., there is an overall increase in precarity). The changes are of different magnitude (and we
will explore the reasons for that next), but it is worth noting that
\emph{income} shocks affect both net worth and health indices because of the
interconnected nature of these attributes in reality.

The health index precarity changes in a less consistent manner: indeed in the
IFP model it appears that health precarity appears to decrease in certain parts
of the distribution. We suspect this is because of two factors: firstly, the
health index is computed relative to the average income level in a particular
state. Thus, even if income decreases, the health index might appear to be
``further'' from that mean value and spuriously indicate a better health index
(see the discussion in Section~\ref{sec:simul-based-meth}). A second cause of
this effect could also be that individuals starting off with high precarity
might have their precarity \emph{reduce} as they see similar states (even if
they are inferior states): this is linked to the way in which the different
terms in the precarity index are weighted.

\subsection{Heterogeneity in precarity evolution}
\label{sec:heter-prec-evol}

\begin{figure*}[!htbp]
    \vspace{-1.0mm}
    \centering
    \begin{subfigure}[t]{0.3\textwidth}
        \centering
        \includegraphics[width=\columnwidth]{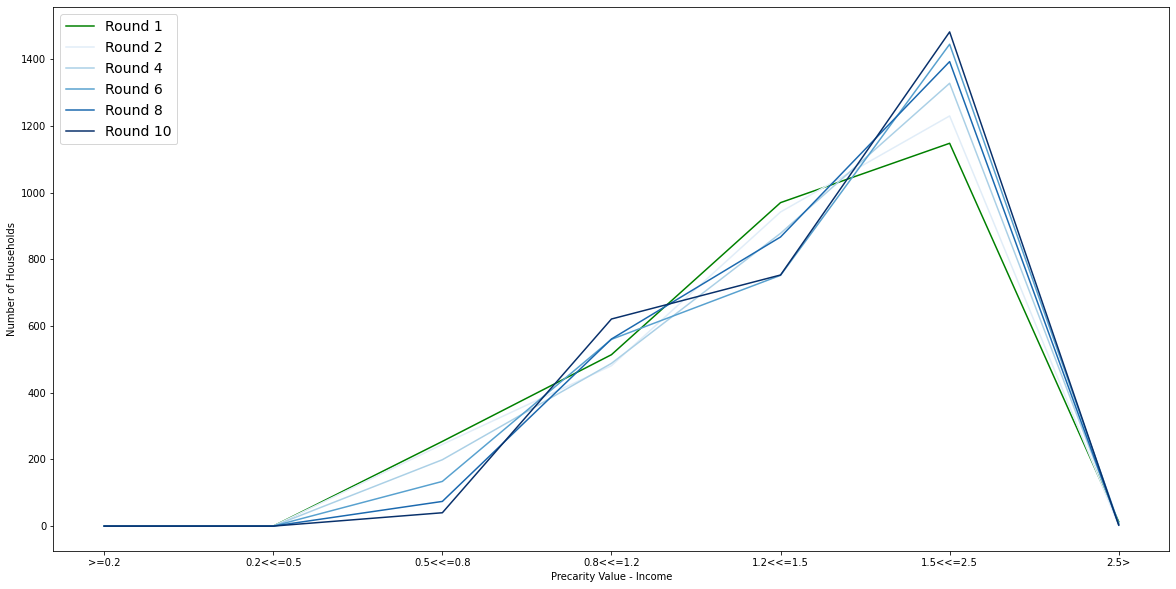}
        \caption{Low income precarity}
        \label{fig:ifp-prec_income_low}
    \end{subfigure}\qquad
    \begin{subfigure}[t]{0.3\textwidth}
        \centering
        \includegraphics[width=\columnwidth]{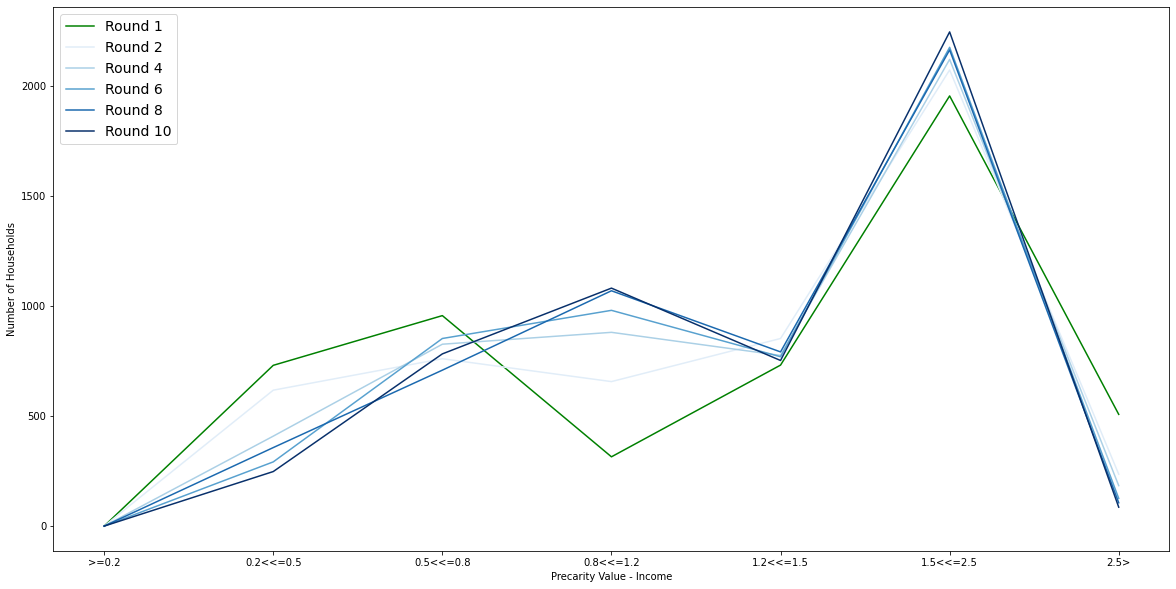}
        \caption{Middle income precarity}
        \label{fig:ifp-prec_income_middle}
    \end{subfigure}\qquad
    \begin{subfigure}[t]{0.3\textwidth}
        \centering
        \includegraphics[width=\columnwidth]{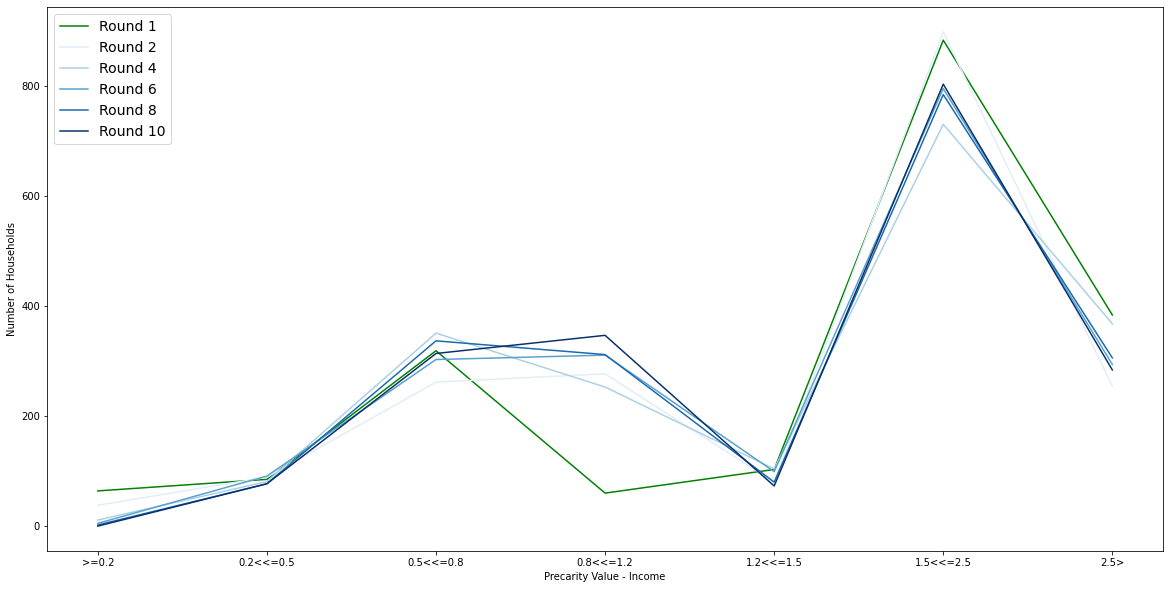}
        \caption{High income precarity}
        \label{fig:ifp-prec_income_high}
    \end{subfigure}

    \caption{Assessing income classes' precarity over time - IFP model}
    \label{fig:ifp income classes}
        
\end{figure*}

\begin{figure*}[!htbp]
    \vspace{-1.0mm}
    \centering
    \begin{subfigure}[t]{0.3\textwidth}
        \centering
        \includegraphics[width=\columnwidth]{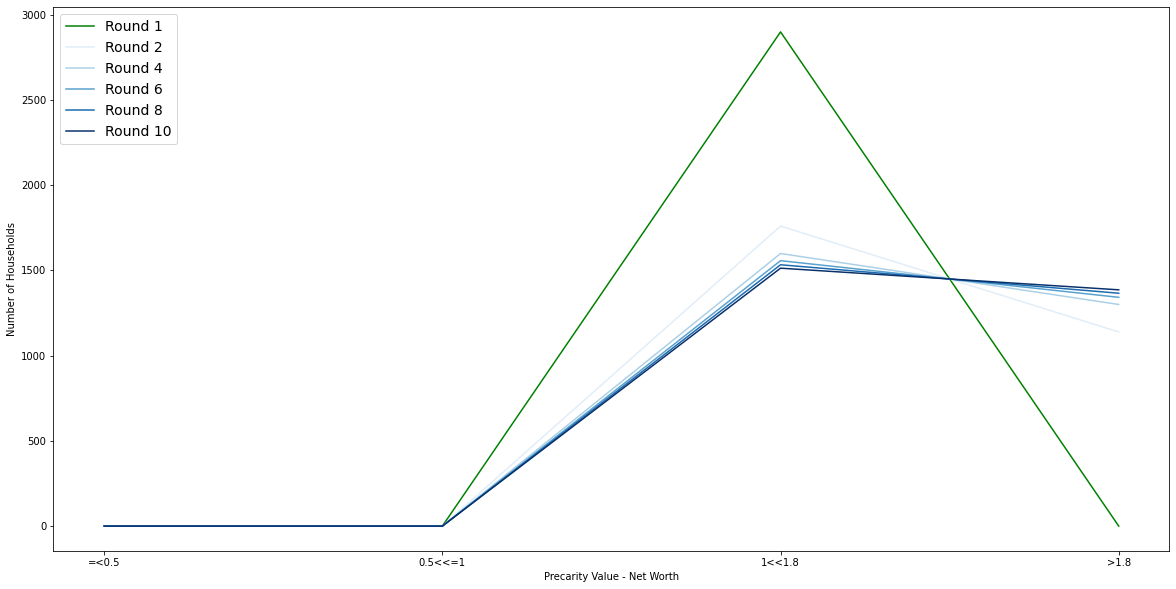}
        \caption{Low income net worth precarity}
        \label{fig:ifp-prec_sav_low}
    \end{subfigure}\qquad
    \begin{subfigure}[t]{0.3\textwidth}
        \centering
        \includegraphics[width=\columnwidth]{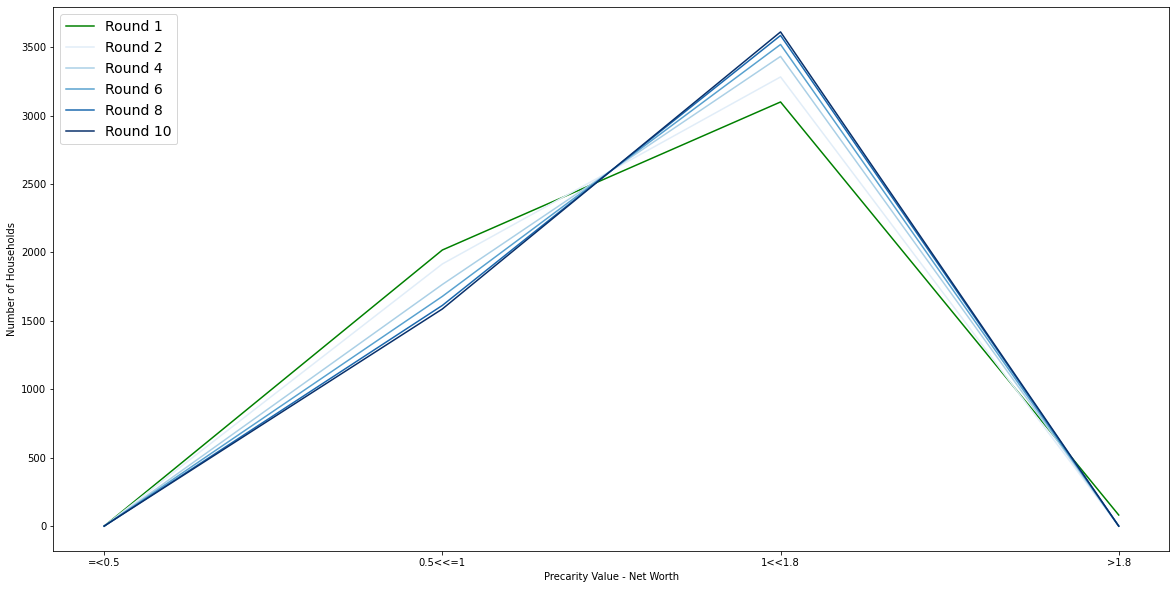}
        \caption{Middle income net worth precarity}
        \label{fig:ifp-prec_sav_mid}
    \end{subfigure}\qquad
    \begin{subfigure}[t]{0.3\textwidth}
        \centering
        \includegraphics[width=\columnwidth]{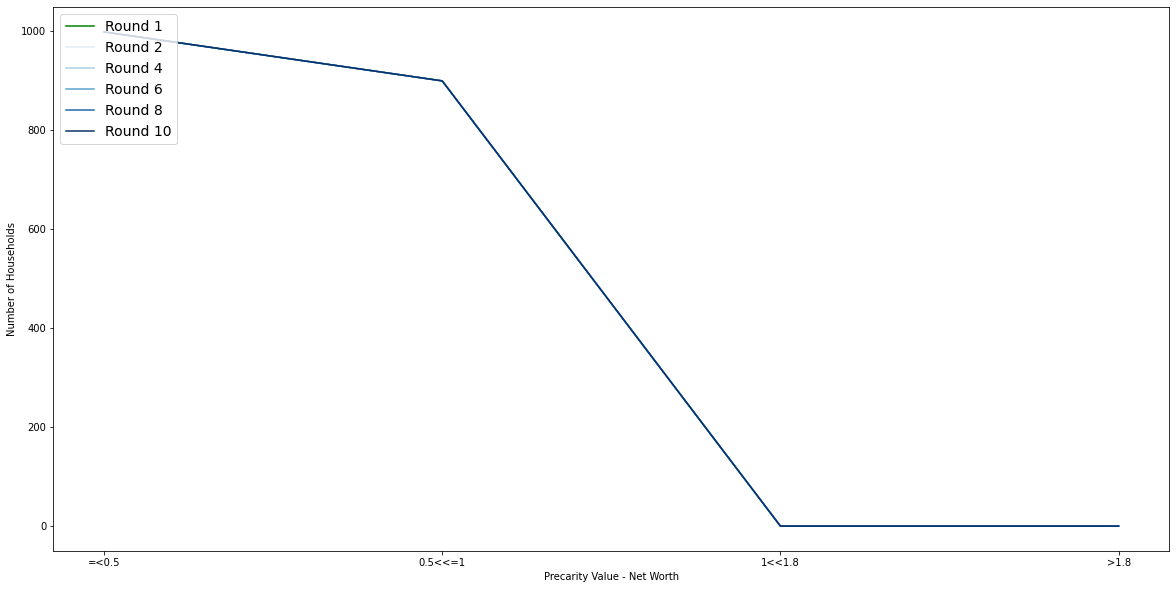}
        \caption{High income class net worth precarity}
        \label{fig:ifp-prec_sav_high}
    \end{subfigure}

    \caption{Assessing income classes' net worth precarity over time - IFP model}
    \label{fig:ifp savings income classes}
        
\end{figure*}

\begin{figure*}[!htbp]
    \vspace{-1.0mm}
    \centering
    \begin{subfigure}[t]{0.3\textwidth}
        \centering
        \includegraphics[width=\columnwidth]{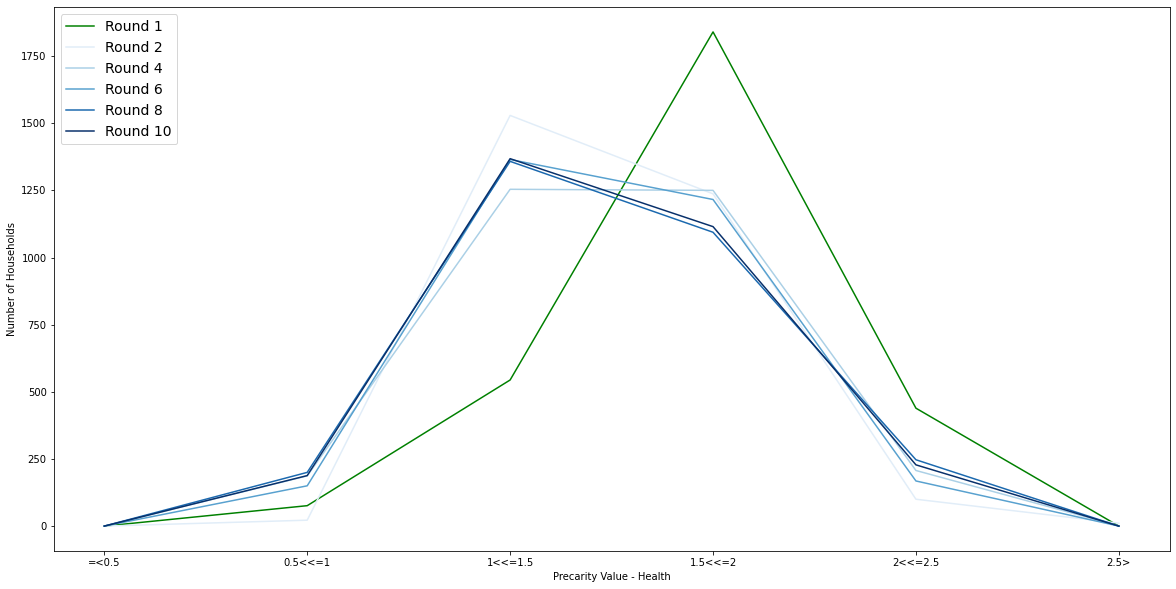}
        \caption{Low income health precarity}
        \label{fig:ifp-prec_health_low}
    \end{subfigure}\qquad
    \begin{subfigure}[t]{0.3\textwidth}
        \centering
        \includegraphics[width=\columnwidth]{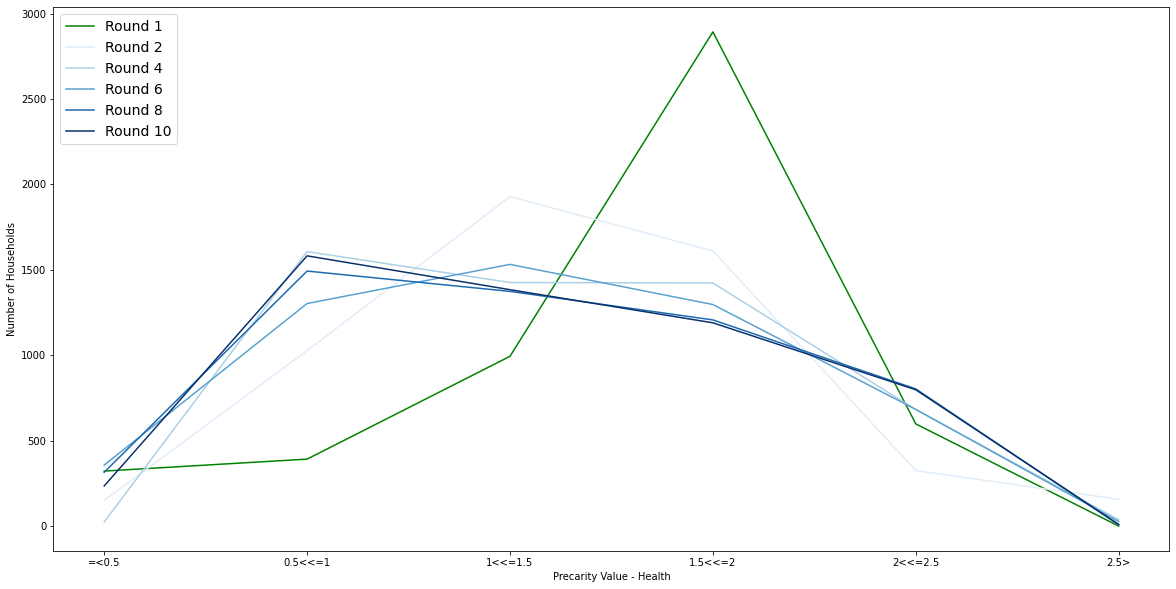}
        \caption{Middle income health precarity}
        \label{fig:ifp-prec_health_mid}
    \end{subfigure}\qquad
    \begin{subfigure}[t]{0.3\textwidth}
        \centering
        \includegraphics[width=\columnwidth]{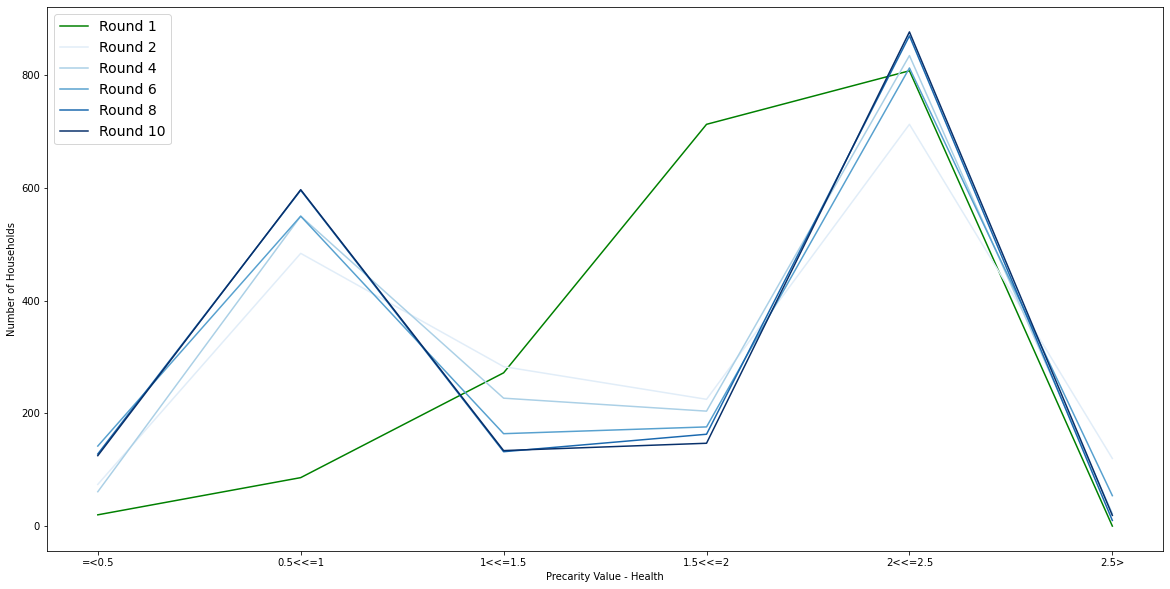}
        \caption{High income class health precarity}
        \label{fig:ifp-prec_health_high}
    \end{subfigure}

    \caption{Assessing income classes' health precarity over time - IFP Model}
    \label{fig:ifp health income classes}
        
\end{figure*}

\begin{figure*}[!htbp]
    \vspace{-1.0mm}
    \centering
    \begin{subfigure}[t]{0.3\textwidth}
        \centering
        \includegraphics[width=\columnwidth]{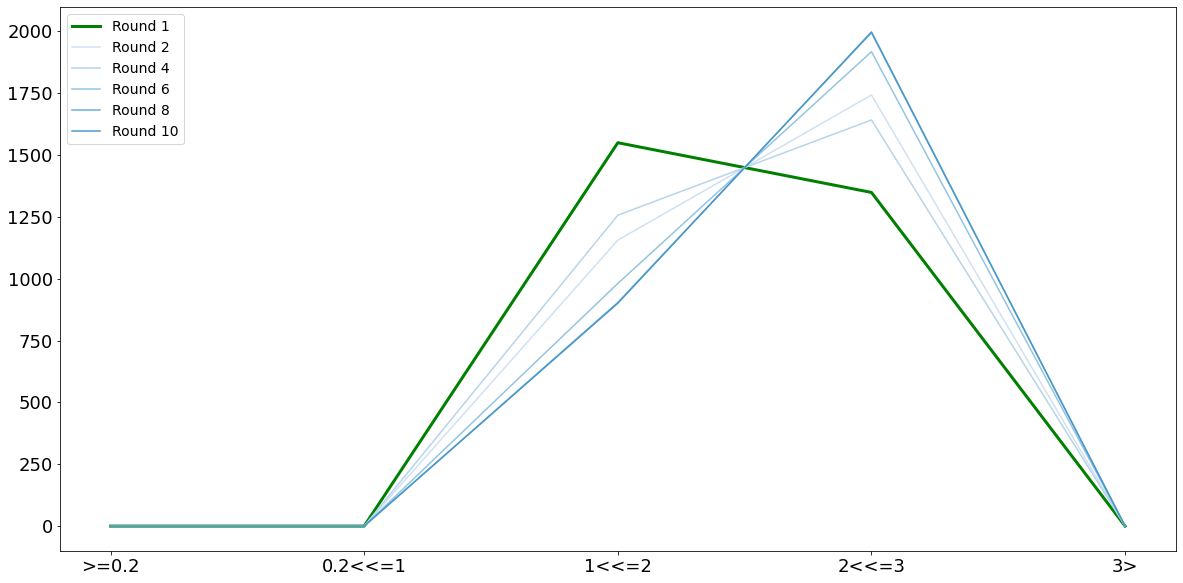}
        \caption{Low income precarity}
        \label{fig:prec_income_low}
    \end{subfigure}\qquad
    \begin{subfigure}[t]{0.3\textwidth}
        \centering
        \includegraphics[width=\columnwidth]{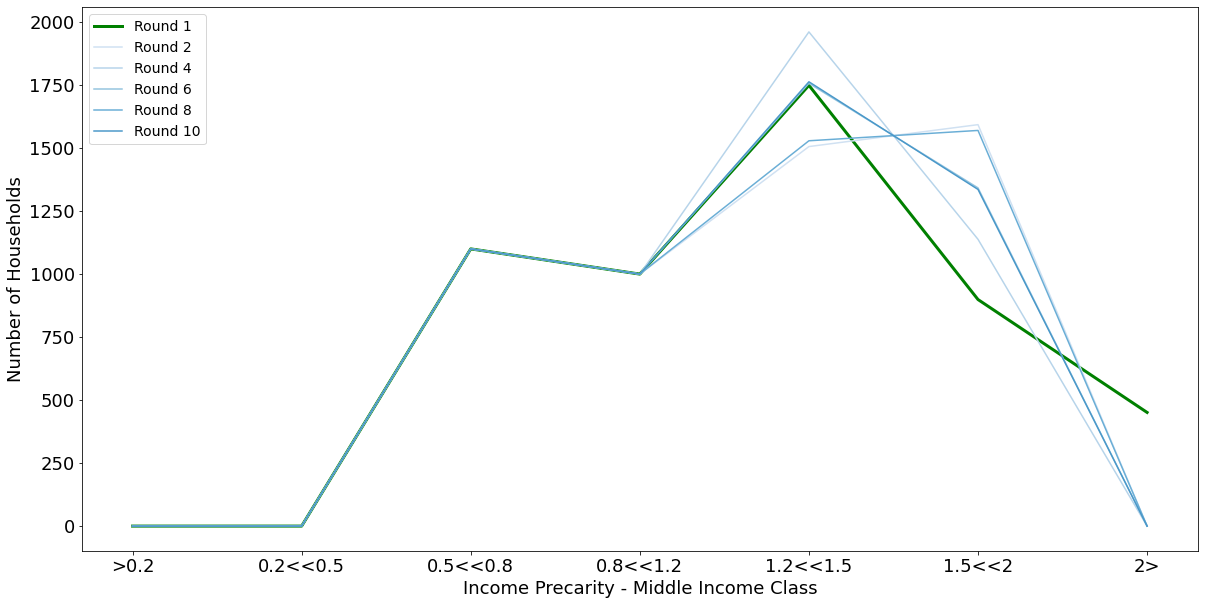}
        \caption{Middle income precarity}
        \label{fig:prec_income_middle}
    \end{subfigure}\qquad
    \begin{subfigure}[t]{0.3\textwidth}
        \centering
        \includegraphics[width=\columnwidth]{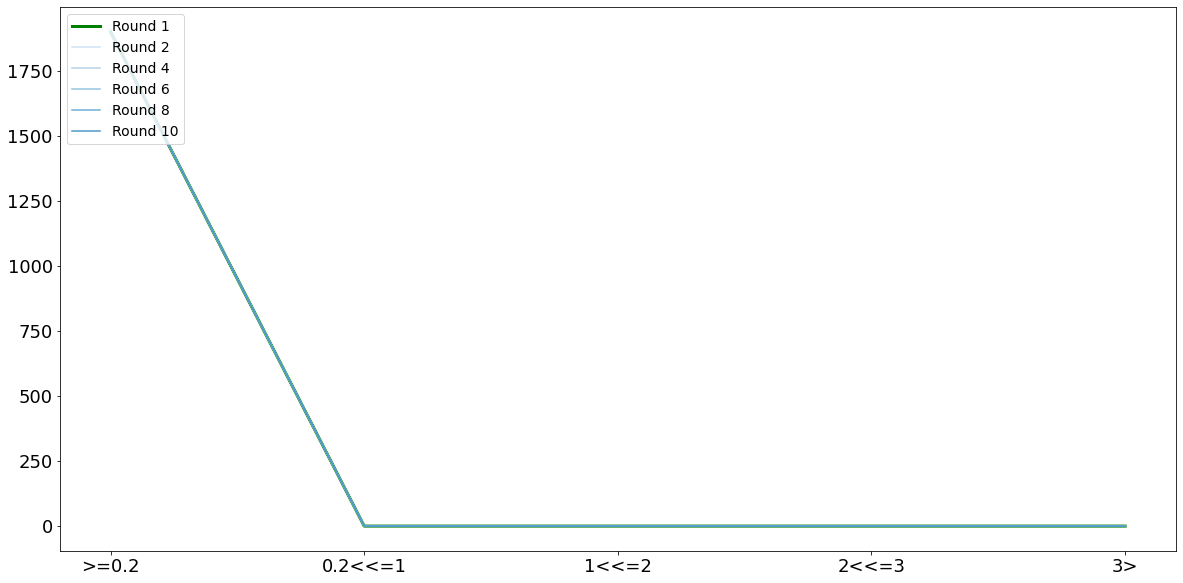}
        \caption{High income precarity}
        \label{fig:prec_income_high}
    \end{subfigure}
    % \caption{Initial health and income distributions}
    % \label{fig:10 rounds precarity}
 
    \caption{Assessing income classes' precarity over time - MDP Model}
    \label{fig:income classes}
        
\end{figure*}

\begin{figure*}[!htbp]
    \vspace{-1.0mm}
    \centering
    \begin{subfigure}[t]{0.3\textwidth}
        \centering
        \includegraphics[width=\columnwidth]{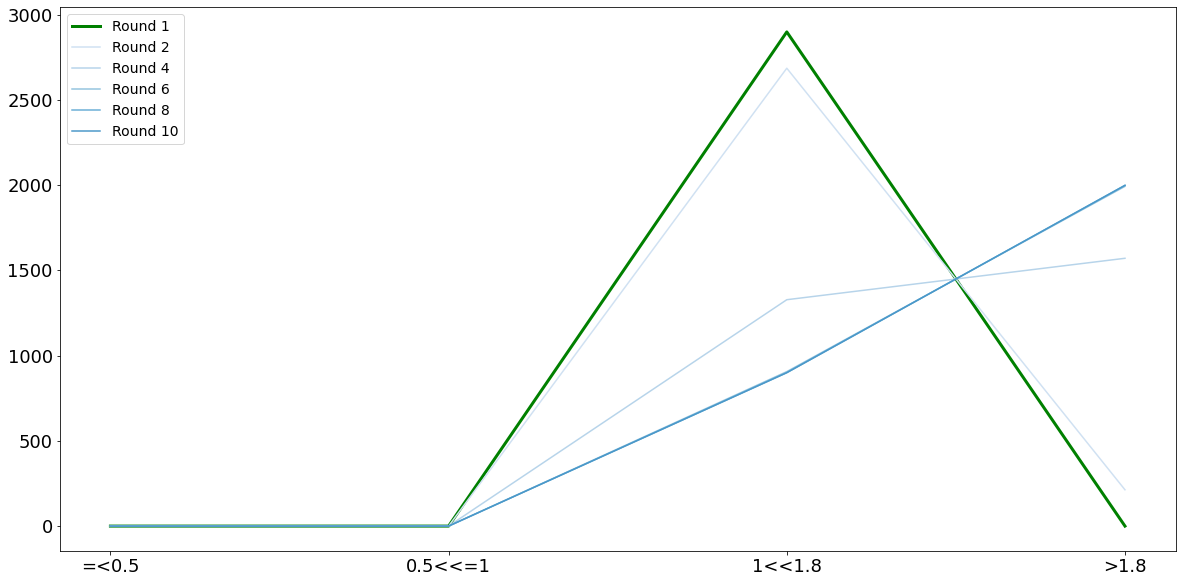}
        \caption{Low income net worth precarity}
        \label{fig:prec_sav_low}
    \end{subfigure}\qquad
    \begin{subfigure}[t]{0.3\textwidth}
        \centering
        \includegraphics[width=\columnwidth]{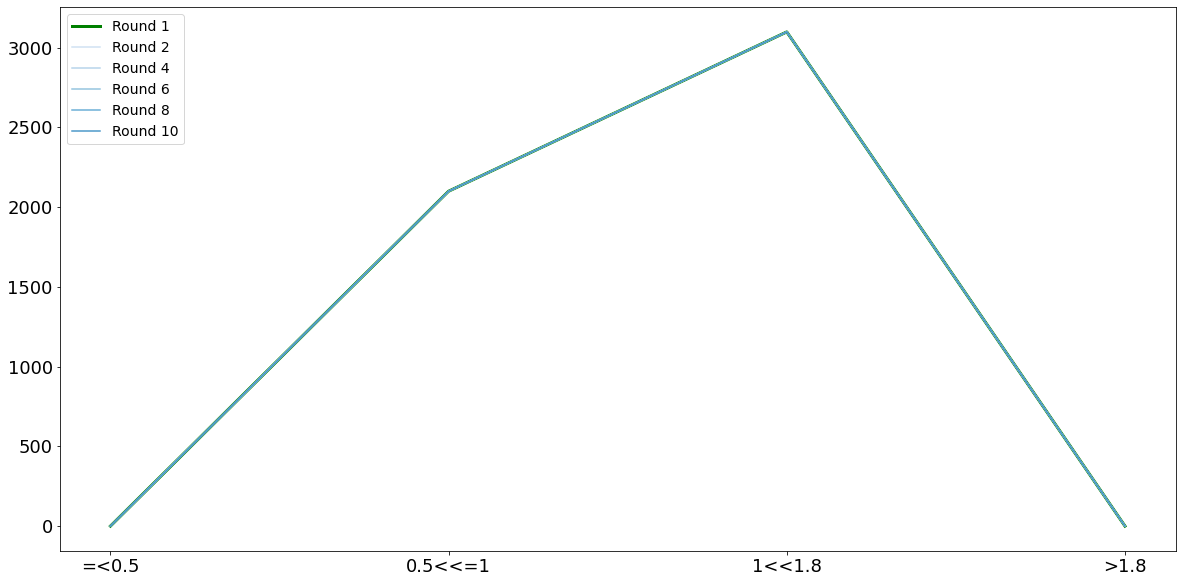}
        \caption{Middle income net worth precarity}
        \label{fig:prec_sav_mid}
    \end{subfigure}\qquad
    \begin{subfigure}[t]{0.3\textwidth}
        \centering
        \includegraphics[width=\columnwidth]{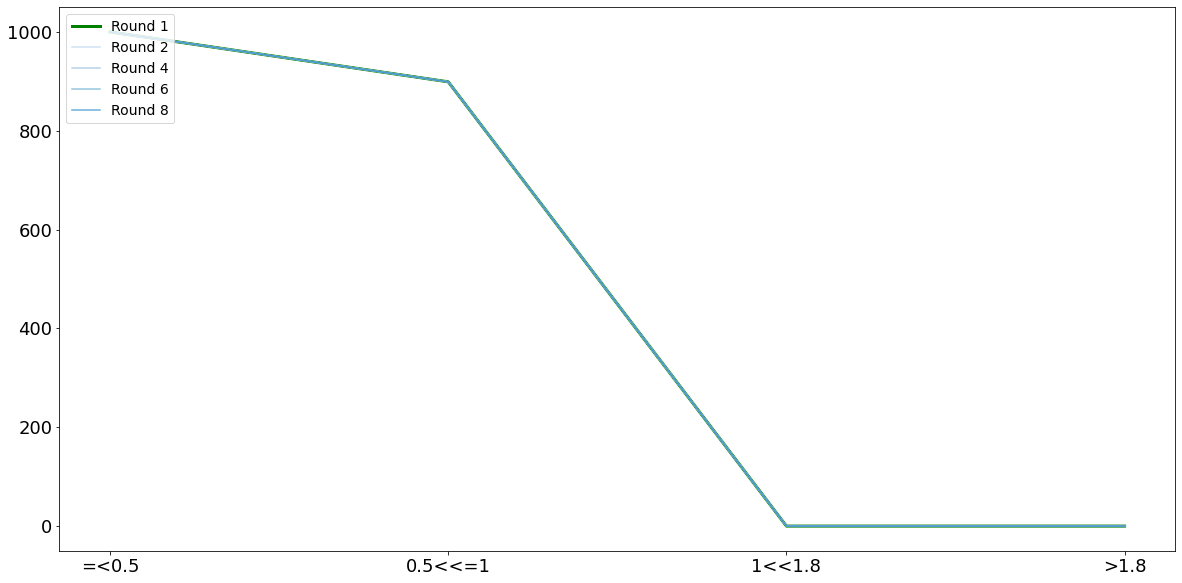}
        \caption{High income net worth precarity}
        \label{fig:prec_sav_high}
    \end{subfigure}
   \caption{Assessing income classes' net worth precarity over time - MDP Model}
    \label{fig:savings income classes}
        
\end{figure*}

\begin{figure*}[!htbp]
    \vspace{-1.0mm}
    \centering
    \begin{subfigure}[t]{0.3\textwidth}
        \centering
        \includegraphics[width=\columnwidth]{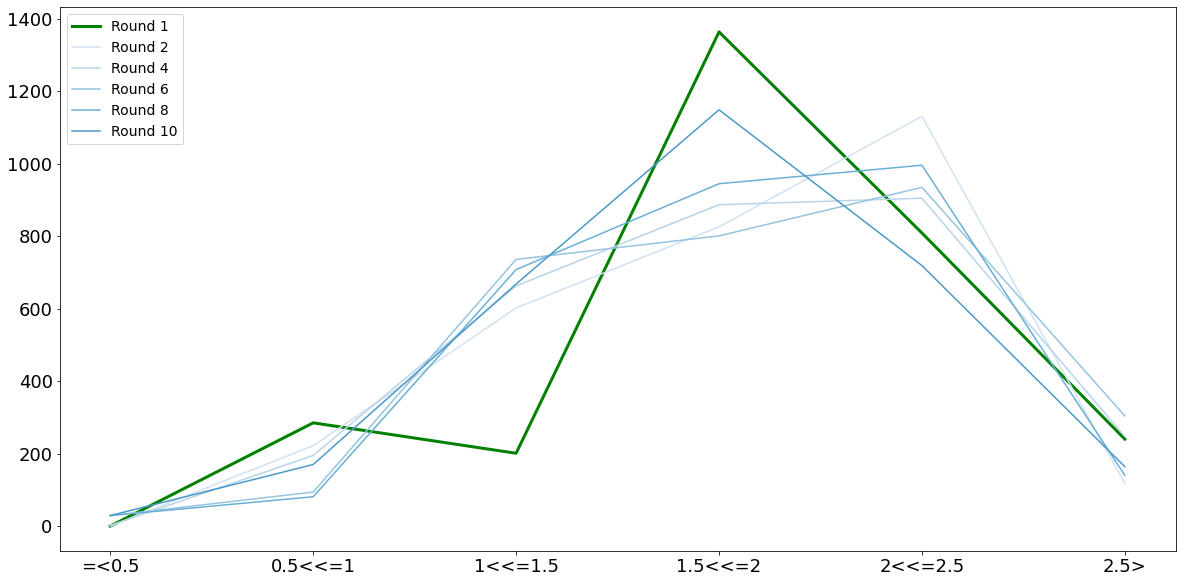}
        \caption{Low income health precarity}
        \label{fig:prec_health_low}
    \end{subfigure}\qquad
    \begin{subfigure}[t]{0.3\textwidth}
        \centering
        \includegraphics[width=\columnwidth]{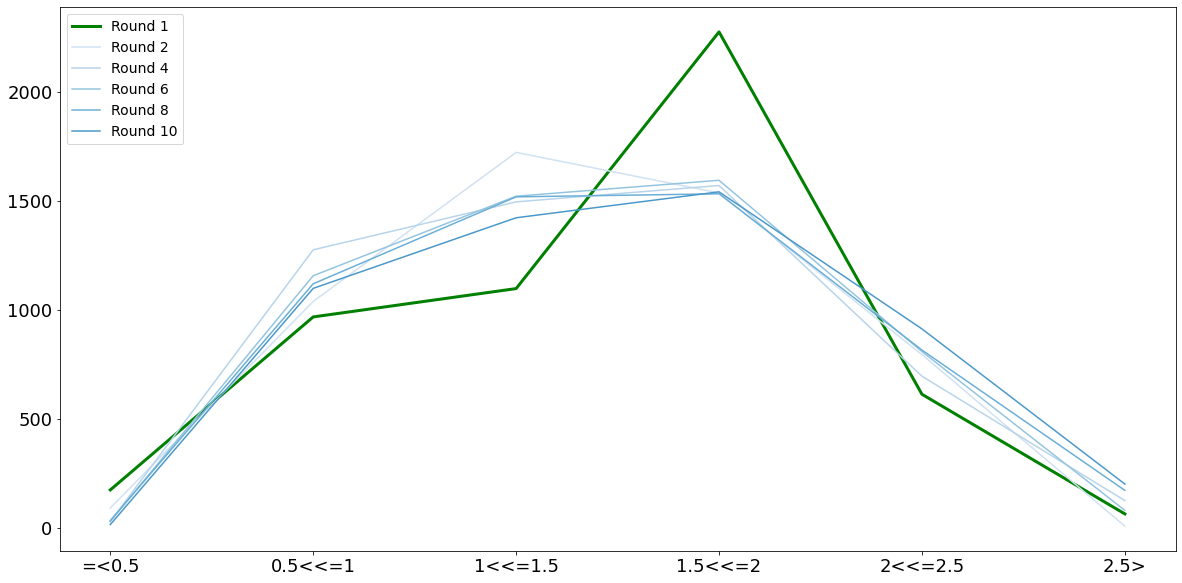}
        \caption{Middle income health precarity}
        \label{fig:prec_health_mid}
    \end{subfigure}\qquad
    \begin{subfigure}[t]{0.3\textwidth}
        \centering
        \includegraphics[width=\columnwidth]{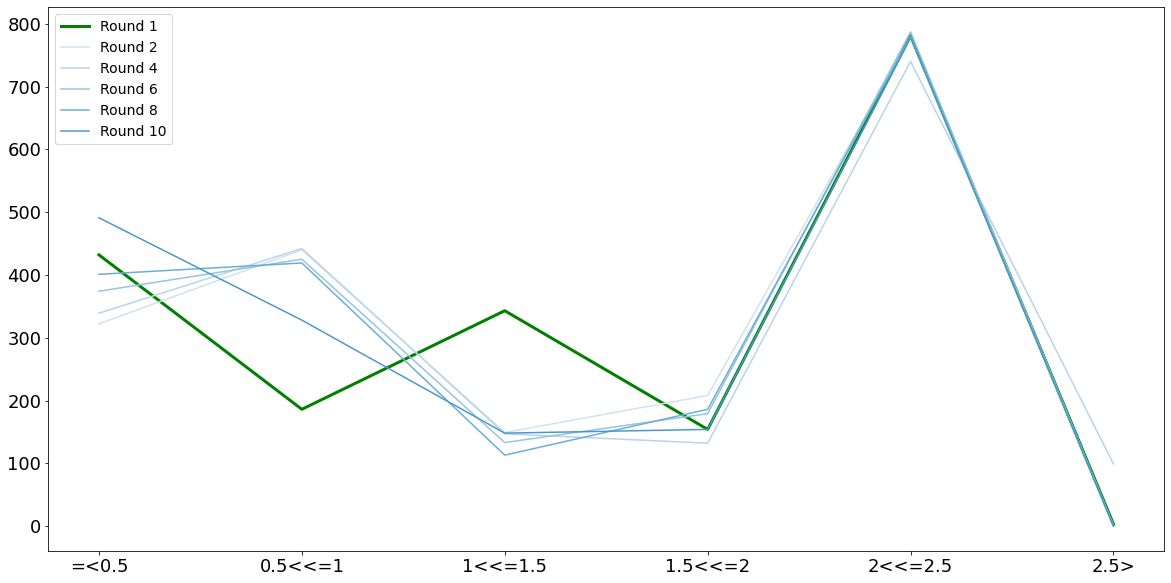}
        \caption{High income health precarity}
        \label{fig:prec_health_high}
    \end{subfigure}
    \caption{Assessing income classes' health precarity over time - MDP Model}
    \label{fig:health income classes}
        
\end{figure*}

The above picture is a global view on precarity across all income levels. One of
the observed effects of precarity is the non-uniform way in which individuals at
different income levels might be affected by financial shocks. To investigate
this, we look at precarity distributions segmented by income level. These
classes are the lower 29\% of the incomes, the middle 52\% of incomes, and the
upper 19\% incomes.\footnote{\url{https://www.pewresearch.org/fact-tank/2020/07/23/are-you-in-the-american-middle-class/}} In the IFP model, the precarity index of income, net worth, and health can be seen in Figures \ref{fig:ifp income classes}, \ref{fig:ifp savings income classes}, and \ref{fig:ifp health income classes}, respectively. Figures \ref{fig:income classes}, \ref{fig:savings income classes} and \ref{fig:health income classes} show this for the MDP model.

\textbf{Analysis.} In general, we see the following consistent behavior. Higher
income individuals maintain a (low) level of precarity over time and sometimes
even experience a \emph{decrease} in precarity. Lower income individuals
experience a clear increase in precarity, and middle income individuals also
experience a precarity increase (but less). In other words, there is a
\emph{compounding} effect of income shocks for individuals who are already in
precarious positions, the exact concern that precarity seeks to capture. While
our simulation is a gross oversimplification of reality, this phenomenon has
been observed in the real world. During the pandemic for example, in March around $34.4\%$ of low income people with income less than \$27,000 lost their job compared to that of only $13.2\%$ high income people with income more than \$60,000.\footnote{\url{https://tracktherecovery.org/}}

\subsection{Policy interventions}
A potential value of a simulation framework is our ability to experiment with
interventions that would be difficult if not impossible to test ``in the
wild''. We demonstrate the value of our simulation with two policy interventions
that might be implemented to alleviate precarity. Both interventions are motivated by concrete measures that have been proposed to
alleviate wealth shocks experienced during the pandemic.

\begin{enumerate}
    \item \textit{Fixed stimulus intervention}: We consider a fixed stimulus
      intervention (measured as a fixed monthly value of \$1500 similar to the
      stimulus monthly checks during the pandemic
      \footnote{\url{https://www.nytimes.com/article/coronavirus-stimulus-package-questions-answers.html}})
      given to all households who fall below the classifier threshold on every
      round. This form of fixed stimulus is similar to the mitigation model
      suggested by Abebe et al.\ \cite{abebe2020subsidy} (although in their model the goal is
      to allocate different fixed amounts of stimulus to different individuals)
  
    \item \textit{Precarity resistance}: An alternate approach to dealing with
      income shocks was demonstrated (among others) by Germany, where the
      government instituted a program to help people keep their jobs and
      continue to be on the payroll.\footnote{Germany's Kurzarbeit Program:
        \url{https://tinyurl.com/yd9qpahs}} We modeled this by reducing the
      probability of a transition to a poorer economic state after an adverse
      decision in our simulations.  We implement this in the MDP model by
      adjusting the transition probabilities directly and in the IFP model by
      adjusting the transition process that generates the exogenous state $Z$. 
\end{enumerate}

\begin{figure*}[!htbp]
    \vspace{-1.0mm}
    \centering
    \begin{subfigure}[t]{0.3\textwidth}
        \centering
        \includegraphics[width=\columnwidth]{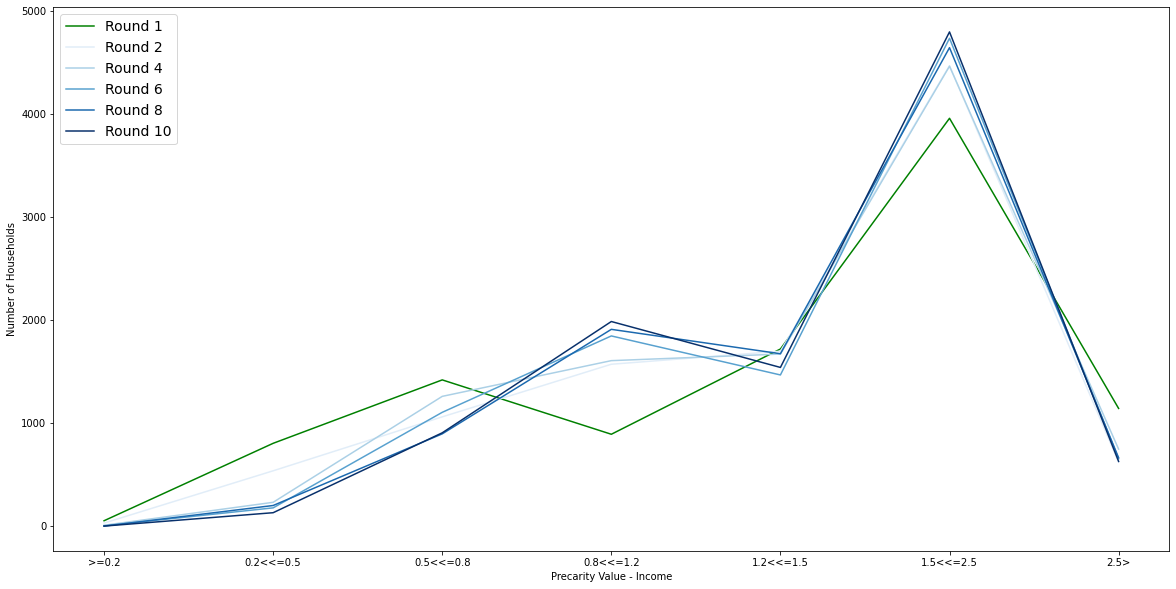}
        \caption{Stimulus effect on income}
        \label{fig:ifp-stimulus income}
    \end{subfigure}\qquad
    \begin{subfigure}[t]{0.3\textwidth}
        \centering
        \includegraphics[width=\columnwidth]{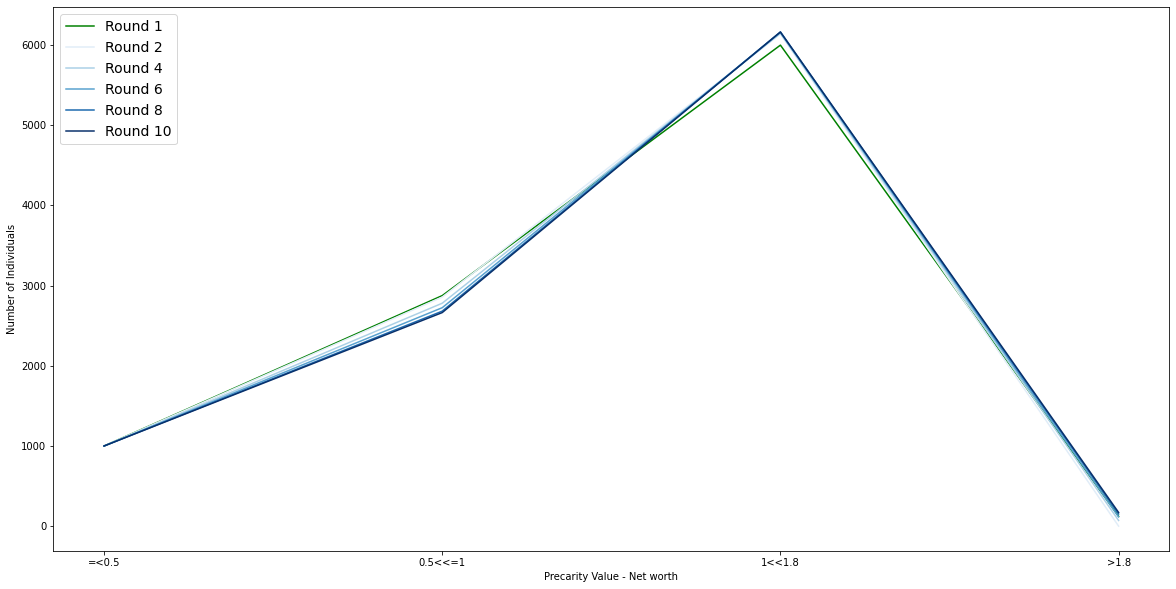}
        \caption{Stimulus effect on net worth}
        \label{fig:ifp-stimulus net worth}
    \end{subfigure}\qquad
    \begin{subfigure}[t]{0.3\textwidth}
        \centering
        \includegraphics[width=\columnwidth]{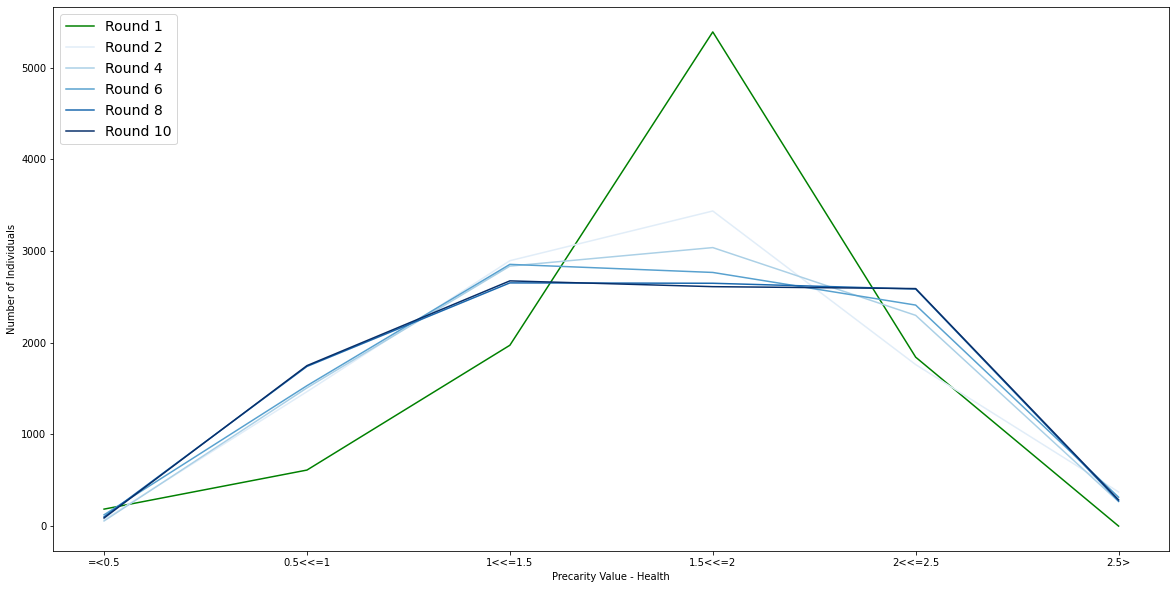}
        \caption{Stimulus effect on health}
        \label{fig:ifp-stimulus health}
    \end{subfigure}%
    
    \caption{Stimulus effects - IFP Model}
    \label{fig:ifp-stimulus}
        
\end{figure*}

\begin{figure*}[!htbp]
    \vspace{-1.0mm}
    \centering
    \begin{subfigure}[t]{0.3\textwidth}
        \centering
        \includegraphics[width=\columnwidth]{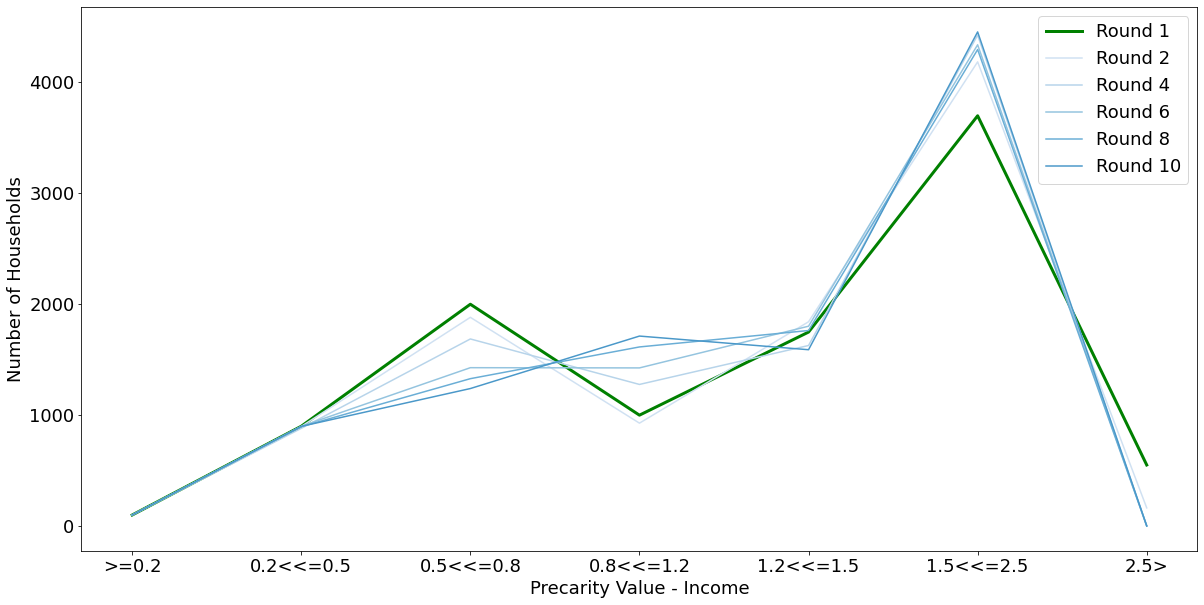}
        \caption{Stimulus effect on income}
        \label{fig:stimulus income}
    \end{subfigure}\qquad
    \begin{subfigure}[t]{0.3\textwidth}
        \centering
        \includegraphics[width=\columnwidth]{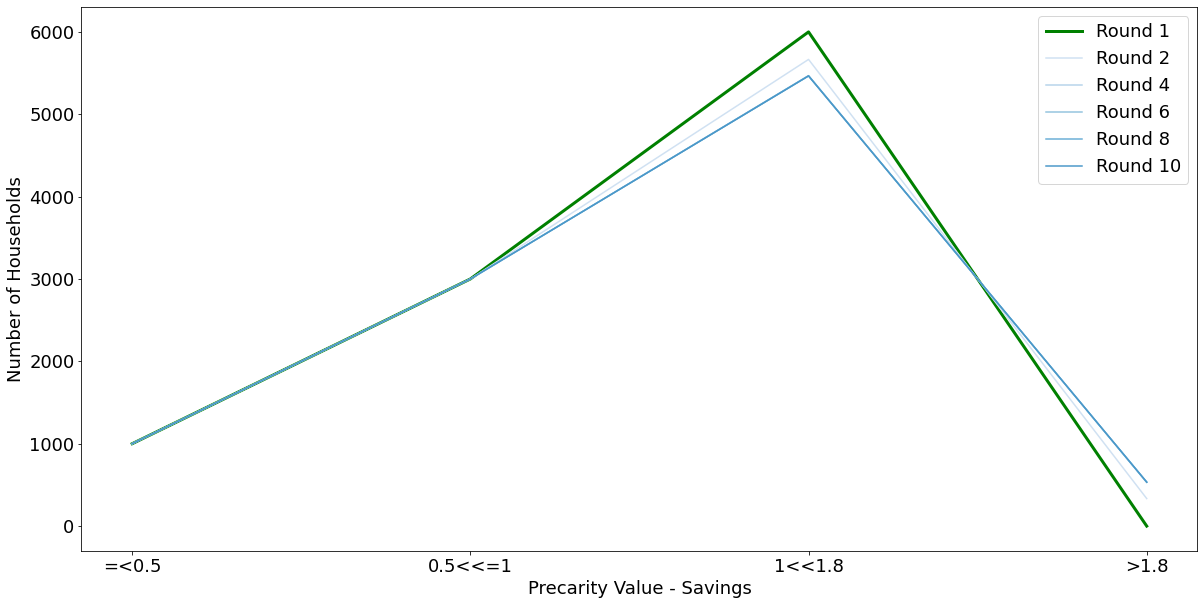}
        \caption{Stimulus effect on net worth}
        \label{fig:stimulus net worth}
    \end{subfigure}\qquad
    \begin{subfigure}[t]{0.3\textwidth}
        \centering
        \includegraphics[width=\columnwidth]{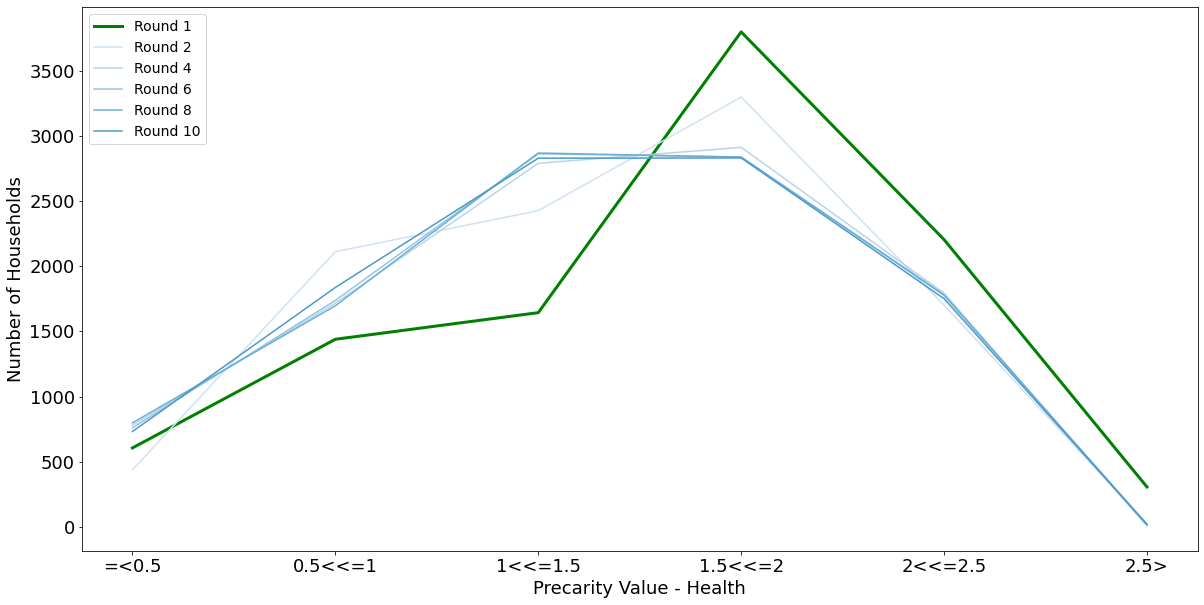}
        \caption{Stimulus effect on health}
        \label{fig:stimulus health}
    \end{subfigure}%
    
    \caption{Stimulus effects - MDP Model}
    \label{fig:stimulus}
        
\end{figure*}

\begin{figure*}[!htbp]
    \vspace{-1.0mm}
    \centering
    \begin{subfigure}[t]{0.3\textwidth}
        \centering
        \includegraphics[width=\columnwidth]{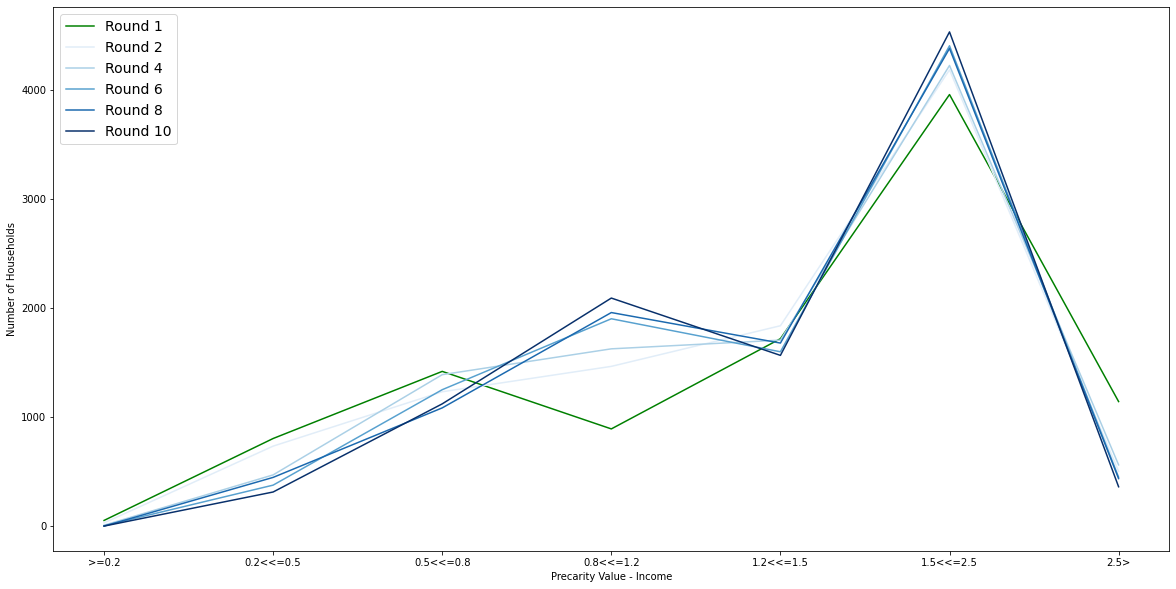}
        \caption{Markov persistence effect on income}
        \label{fig:ifp-markov income}
    \end{subfigure}\qquad
    \begin{subfigure}[t]{0.3\textwidth}
        \centering
        \includegraphics[width=\columnwidth]{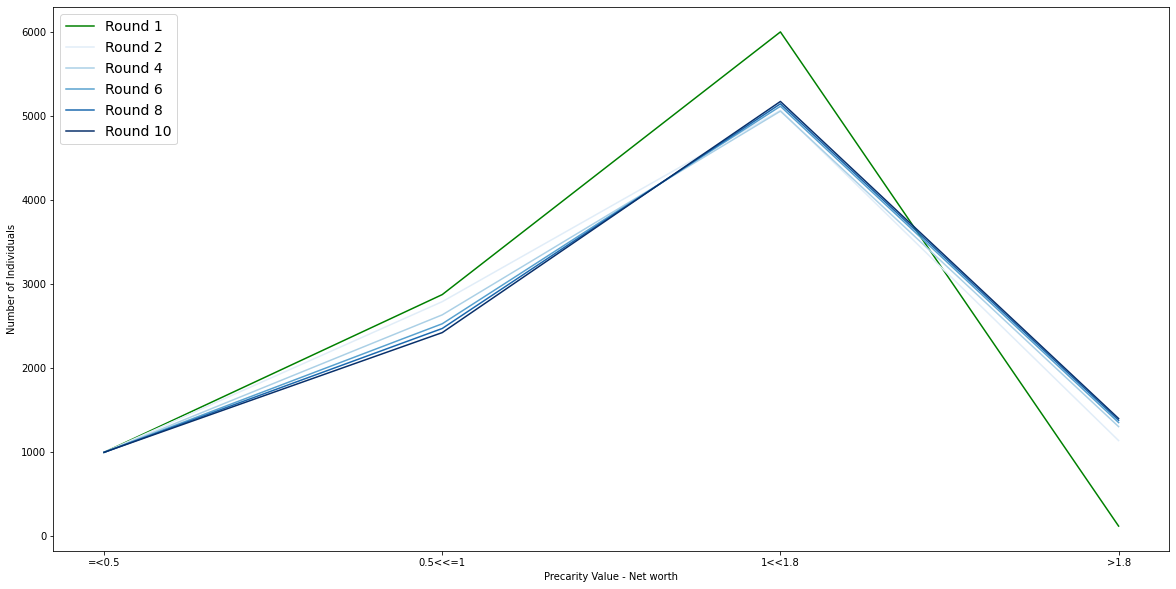}
        \caption{Markov persistence effect on net worth}
        \label{fig:ifp-markov net worth}
    \end{subfigure}\qquad
    \begin{subfigure}[t]{0.3\textwidth}
        \centering
        \includegraphics[width=\columnwidth]{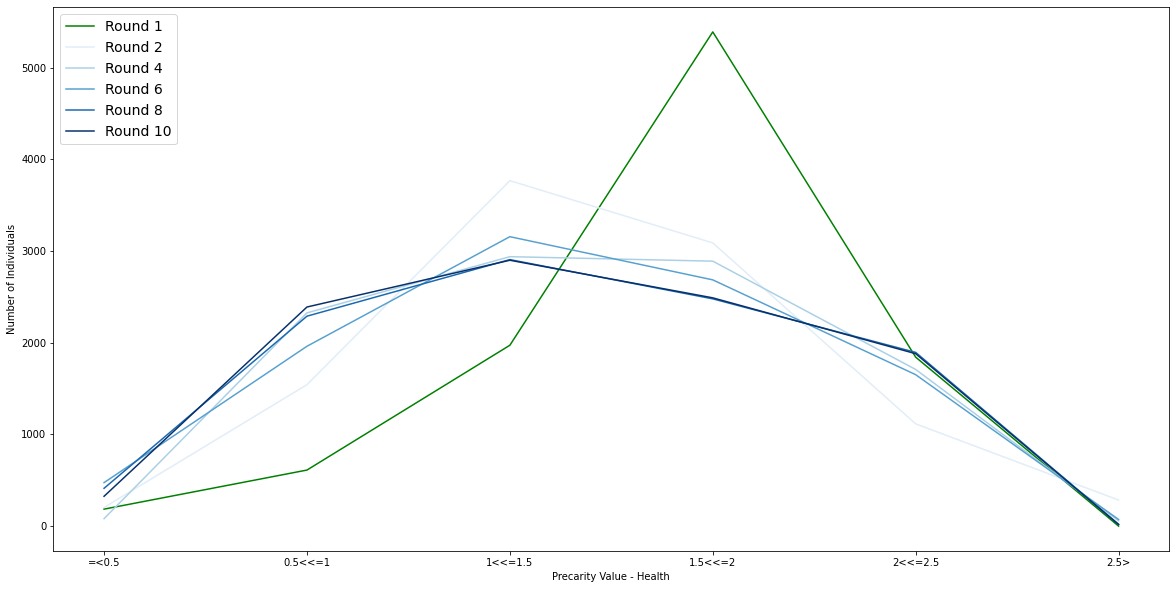}
        \caption{Markov persistence effect on health}
        \label{fig:ifp-markov health}
    \end{subfigure}%
    
    \caption{Precarity resistance - IFP Model}
    \label{fig:ifp-markov}
\end{figure*}

\begin{figure*}[!htbp]
    \vspace{-1.0mm}
    \centering
    \begin{subfigure}[t]{0.3\textwidth}
        \centering
        \includegraphics[width=\columnwidth]{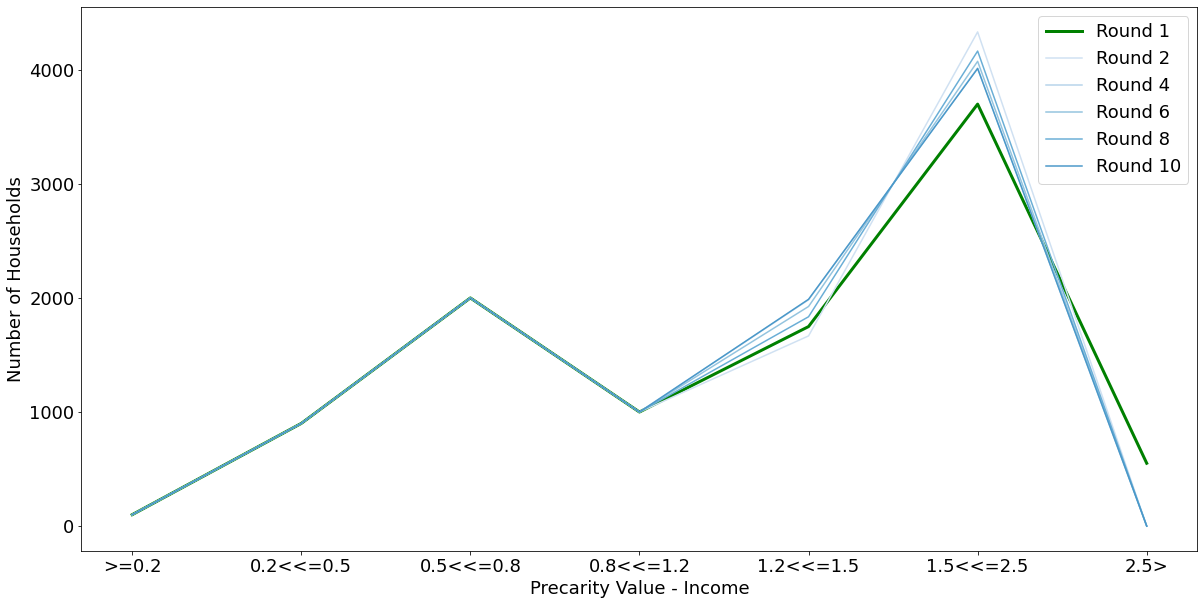}
        \caption{Markov persistence effect on income}
        \label{fig:markov income}
    \end{subfigure}\qquad
    \begin{subfigure}[t]{0.3\textwidth}
        \centering
        \includegraphics[width=\columnwidth]{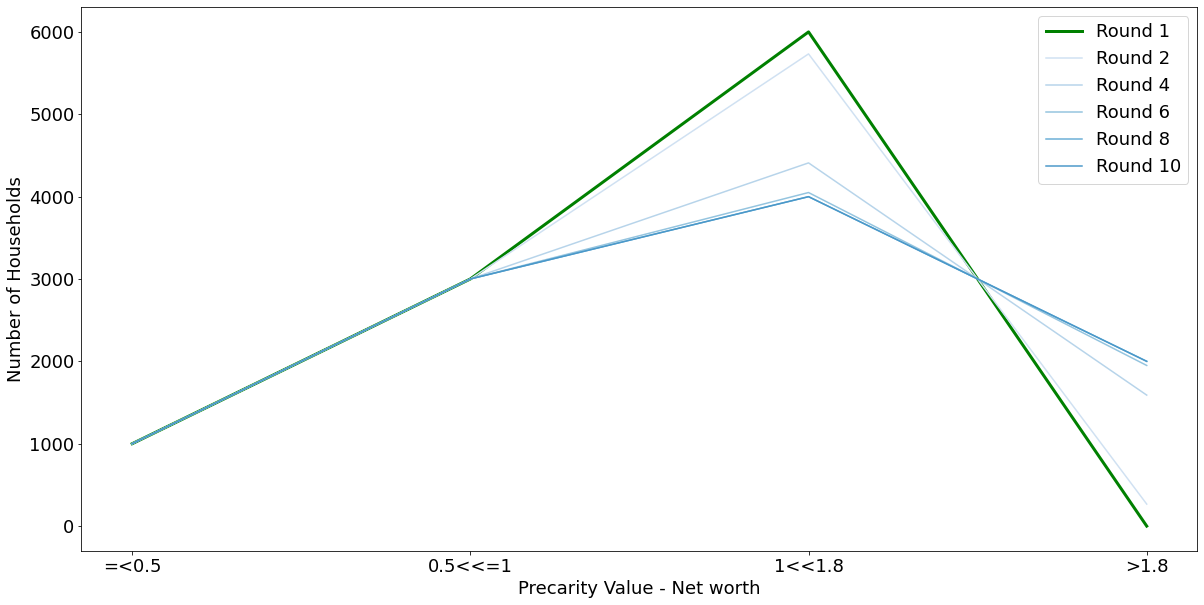}
        \caption{Markov persistence effect on net worth}
        \label{fig:markov net worth}
    \end{subfigure}\qquad
    \begin{subfigure}[t]{0.3\textwidth}
        \centering
        \includegraphics[width=\columnwidth]{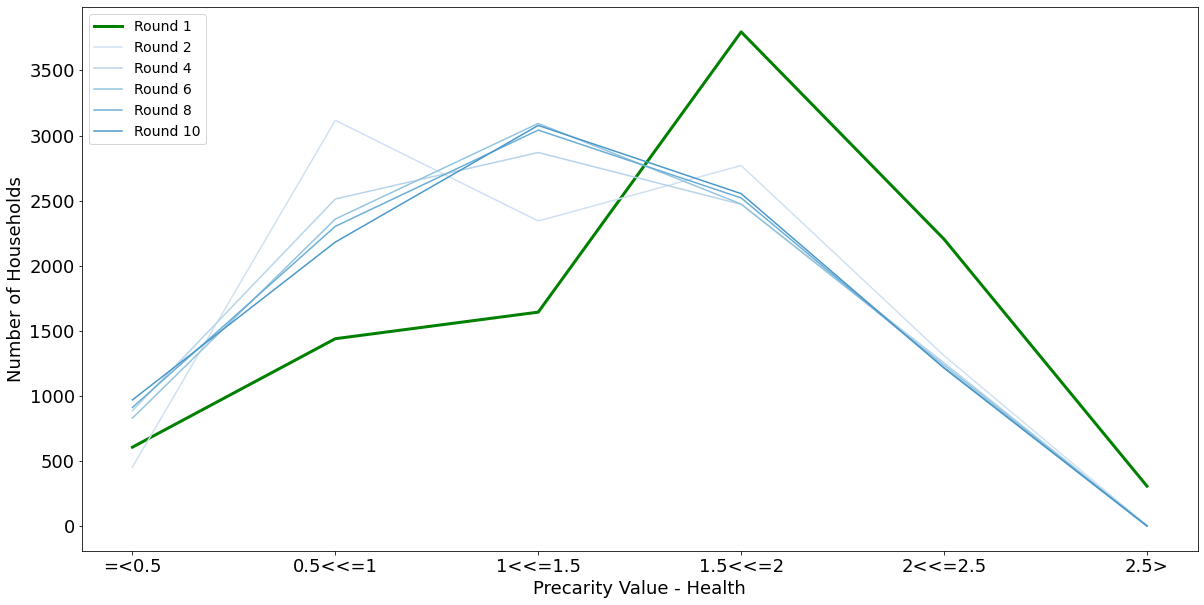}
        \caption{Markov persistence effect on health}
        \label{fig:markov health}
    \end{subfigure}%
    
    \caption{Precarity resistance - MDP Model}
    \label{fig:markov}
        
\end{figure*}

We show the out-turn of the same policy interventions in the IFP model. These results are shown in Figures  \ref{fig:ifp-stimulus} and
\ref{fig:ifp-markov}. Figures \ref{fig:stimulus} and \ref{fig:markov} show the
results for the MDP model, respectively.

\textbf{Analysis.} We see that these interventions have a measurable effect on decreasing household
precarity compared to Figure \ref{fig:10 rounds precarity}, as the number of
households with higher precarity indices reduces. The effect of enforcing an
intervention on income is of a ripple effect on other tied
features: we also observe a decline in precarity for these features.  In
addition, we see a measurable effect of the stimulus on the net worth and the
count of households in the lowest income precarity values.

Note that the changes in the IFP model are more subtle because in the  IFP model
a household consumes as much as possible, constrained by utility and basic
needs. Therefore, although assets increase with the \$1500 stimulus, it also
increases the consumption. Note that we employed the interventions from the
first round onward. We also tested the effects with interventions in later
rounds (e.g., round 6 onward). The effects of such interventions is negligible,
implying that reacting to the underlying population's precarity after they are
precarious beyond the point of recovery would have little to no effect.

\section{Discussion}

The main contribution of this work is the introduction of the rich sociological
and the economic notion of precarity to the community of researchers thinking about
automated decision making, a simulation framework to experiment with it and an
empirical study of how precarity manifests in a simplified
macroeconomic system.

We believe that the study of precarity is important for two reasons. Firstly, it
takes the focus of decision making away from the decision \emph{maker} and their
goals for maximizing utility and other socially desirable goals, and towards the
experience of an individual subject to a sequence of decisions. Secondly, this ``averaging
over time'' reveals phenomena of inequality that are hidden underneath gross
population-wide measures of progress. Lastly, although this work is noted to the Artificial Intelligence community, the contributions are also important for public policy. 

\paragraph{Group inequality and precarity:}
Thus far, all simulations use parameters that apply to the entire population of households.  While individuals in the simulation start at different states, incomes, wealth, and health, the model is identical for every individual.  Yet, it is clear that different demographic groups experience bias and discrimination in society.  Historical discrimination has a huge impact on an individual's starting point \cite{baradaran2019color, abbasi2019fairness, gupta2019equalizing}, and ongoing discrimination has a major impact on one's income even for the same job (via gender \cite{blau2017gender} or ability \cite{whittaker2019disability}), income shocks \cite{zwerling1992race}, health shocks \cite{hicken2014racial}, and wealth shocks, as well as the likelihood of benefiting from public policy.  In a sense, one's demographic group changes how the economic model must be run.  One might apply this model to members of a particular intersectional group who benefit from or are harmed by discrimination in the same manner.  Future work could also use multiple models, one each for individuals in different intersectional groups, to show how model differences can lead to disparate results by groups in precarity and financial position in society, for example, why the US wealth gap by race has remained largely unchanged over many decades \cite{baradaran2019color}.

\paragraph{Limitations:}

Our framework, like any simulation framework, is limited by what we have chosen
to retain and omit. Our simulation assumes societal homogeneity -- all
individuals are subject to the same forces and constraints, outside of economic
differences. This of course, fails to reflect other forms of inequity in society
like the ones described above, and is a direction for further exploration. The system could, however, if there is heterogeneity in individuals, take that into account in its decisions.

Our choice of income, net worth and health as state variables is well-grounded
in the literature on macroeconomic models. However, our modeling of how income
and wealth factors influence and are influenced by health relies on a simple
formula equating these factors.

Our two choices of simulation frameworks reflect two extreme perspectives in
economic modeling: the infinite-horizon optimizer and the myopic local decision
maker. The truth usually lies somewhere in between, and we expect that a more
sophisticated modeling of bounded rationality might allow us to capture a more
nuanced and realistic set of behaviors.

Finally, while our model behavior and predictions broadly align with observed
patterns of inequality that we see right now, a more reliable validation
would be to see if we can match trends in income and wealth distributions more
precisely. It would also be instructive to develop strong tests of
significance for the shifts that we observe in response to interventions.

\section*{Acknowledgements}
\label{sec:aknowledgement}

We thank Vivek Gupta for his valuable insights at various stages of the project. We also thank anonymous reviewers for their helpful suggestions and comments.

%%% Local Variables:
%%% mode: latex
%%% TeX-master: "main"
%%% End:

\cleardoublepage
\bibliographystyle{unsrt}  
\bibliography{references}

\begin{thebibliography}{10}

\bibitem{anthro}
Jess Benhabib and Alberto Bisin.
\newblock {Precarity}.
\newblock {\em Sharryn Kasmir, The Cambridge Encyclopedia of Anthropology},
  2018.

\bibitem{nancyworth}
Nancy Worth.
\newblock {Making sense of precarity: talking about economic insecurity with
  millennials in Canada}.
\newblock {\em Journal of Cultural Economy}, 12(5):441--447, 2019.

\bibitem{greigDepeuter}
Greig de~Peuter.
\newblock {Creative Economy and Labor Precarity: A Contested Convergence}.
\newblock {\em Journal of Communication Inquiry}, 35(4):417--425, 2011.

\bibitem{butler2006precarious}
Judith Butler.
\newblock {\em {Precarious life: The powers of mourning and violence}}.
\newblock Verso Books, 2006.

\bibitem{butler2016frames}
Judith Butler.
\newblock {\em {Frames of war: When is life grievable?}}
\newblock Verso Books, 2016.

\bibitem{standing2014precariat}
Guy Standing.
\newblock {The Precariat-The new dangerous class}.
\newblock {\em Amalgam}, 6(6-7):115--119, 2014.

\bibitem{gill2016century}
Lesley Gill.
\newblock {\em {A century of violence in a Red City: Popular struggle,
  counterinsurgency, and human rights in Colombia}}.
\newblock Duke University Press, 2016.

\bibitem{allison2014precarious}
Anne Allison.
\newblock {\em {Precarious Japan}}.
\newblock Duke University Press, 2014.

\bibitem{ritschard2018index}
Gilbert Ritschard, Margherita Bussi, and Jacqueline O’Reilly.
\newblock {An index of precarity for measuring early employment insecurity}.
\newblock In {\em Sequence Analysis and Related Approaches}, pages 279--295.
  Springer, 2018.

\bibitem{aneja2019no}
Abhay~P Aneja and Carlos~F Avenancio-Le{\'o}n.
\newblock {No Credit For Time Served? Incarceration and Credit-Driven Crime
  Cycles}.
\newblock {\em Mostly Harmless Econometrics. Princeton: Princeton University
  Press.}, 2019.

\bibitem{abebe2020subsidy}
Rediet Abebe, Jon Kleinberg, and S~Matthew Weinberg.
\newblock {Subsidy allocations in the presence of income shocks}.
\newblock In {\em Proceedings of the AAAI Conference on Artificial
  Intelligence}, volume~34, pages 7032--7039, 2020.

\bibitem{zhang2020fairness}
Xueru Zhang and Mingyan Liu.
\newblock {Fairness in Learning-Based Sequential Decision Algorithms: A
  Survey}.
\newblock {\em arXiv preprint arXiv:2001.04861}, 2020.

\bibitem{heidari2018preventing}
Hoda Heidari and Andreas Krause.
\newblock {Preventing Disparate Treatment in Sequential Decision Making.}
\newblock In {\em IJCAI}, pages 2248--2254, 2018.

\bibitem{gupta2019individual}
Swati Gupta and Vijay Kamble.
\newblock Individual fairness in hindsight.
\newblock In {\em Proceedings of the 2019 ACM Conference on Economics and
  Computation}, pages 805--806, 2019.

\bibitem{bechavod2019equal}
Yahav Bechavod, Katrina Ligett, Aaron Roth, Bo~Waggoner, and Steven~Z Wu.
\newblock {Equal opportunity in online classification with partial feedback}.
\newblock In {\em Advances in Neural Information Processing Systems}, pages
  8972--8982, 2019.

\bibitem{joseph2018meritocratic}
Matthew Joseph, Michael Kearns, Jamie Morgenstern, Seth Neel, and Aaron Roth.
\newblock {Meritocratic fairness for infinite and contextual bandits}.
\newblock In {\em Proceedings of the 2018 AAAI/ACM Conference on AI, Ethics,
  and Society}, pages 158--163, 2018.

\bibitem{hebert2017calibration}
Ursula H{\'e}bert-Johnson, Michael~P. Kim, Omer Reingold, and Guy~N. Rothblum.
\newblock {Calibration for the (computationally-identifiable) masses}.
\newblock {\em arXiv preprint arXiv:1711.08513}, 2017.

\bibitem{valera2018enhancing}
Isabel Valera, Adish Singla, and Manuel~Gomez Rodriguez.
\newblock {Enhancing the accuracy and fairness of human decision making}.
\newblock In {\em Advances in Neural Information Processing Systems}, pages
  1769--1778, 2018.

\bibitem{li2019combinatorial}
Fengjiao Li, Jia Liu, and Bo~Ji.
\newblock {Combinatorial sleeping bandits with fairness constraints}.
\newblock {\em IEEE Transactions on Network Science and Engineering}, 2019.

\bibitem{chen2019fair}
Yifang Chen, Alex Cuellar, Haipeng Luo, Jignesh Modi, Heramb Nemlekar, and
  Stefanos Nikolaidis.
\newblock {Fair Contextual Multi-Armed Bandits: Theory and Experiments}.
\newblock {\em arXiv preprint arXiv:1912.08055}, 2019.

\bibitem{gillen2018online}
Stephen Gillen, Christopher Jung, Michael Kearns, and Aaron Roth.
\newblock {Online learning with an unknown fairness metric}.
\newblock In {\em Advances in Neural Information Processing Systems}, pages
  2600--2609, 2018.

\bibitem{patil2019achieving}
Vishakha Patil, Ganesh Ghalme, Vineet Nair, and Y~Narahari.
\newblock {Achieving fairness in the stochastic multi-armed bandit problem}.
\newblock {\em arXiv preprint arXiv:1907.10516}, 2019.

\bibitem{dwork2018fairness}
Cynthia Dwork and Christina Ilvento.
\newblock {Fairness Under Composition}.
\newblock In {\em 10th Innovations in Theoretical Computer Science Conference
  (ITCS 2019)}. Schloss Dagstuhl-Leibniz-Zentrum fuer Informatik, 2018.

\bibitem{liu2018delayed}
Lydia~T Liu, Sarah Dean, Esther Rolf, Max Simchowitz, and Moritz Hardt.
\newblock {Delayed Impact of Fair Machine Learning}.
\newblock In {\em Proceedings of the 35th International Conference on Machine
  Learning}, 2018.

\bibitem{heidari2019long}
Hoda Heidari, Vedant Nanda, and Krishna Gummadi.
\newblock {On the Long-term Impact of Algorithmic Decision Policies: Effort
  Unfairness and Feature Segregation through Social Learning}.
\newblock In {\em International Conference on Machine Learning}, pages
  2692--2701, 2019.

\bibitem{downstream}
Sampath Kannan, Aaron Roth, and Juba Ziani.
\newblock Downstream effects of affirmative action.
\newblock In {\em Proceedings of the Conference on Fairness, Accountability,
  and Transparency}, FAT* ’19, page 240–248, New York, NY, USA, 2019.
  Association for Computing Machinery.

\bibitem{labor_market_lily_hu}
Lily Hu and Yiling Chen.
\newblock A short-term intervention for long-term fairness in the labor market.
\newblock In {\em Proceedings of the 2018 World Wide Web Conference}, WWW
  ’18, page 1389–1398, Republic and Canton of Geneva, CHE, 2018.
  International World Wide Web Conferences Steering Committee.

\bibitem{hashimoto2018fairness}
Tatsunori Hashimoto, Megha Srivastava, Hongseok Namkoong, and Percy Liang.
\newblock {Fairness Without Demographics in Repeated Loss Minimization}.
\newblock In {\em International Conference on Machine Learning}, pages
  1929--1938, 2018.

\bibitem{zhang2019group}
Xueru Zhang, Mohammadmahdi Khaliligarekani, Cem Tekin, and Mingyan Liu.
\newblock {Group Retention when Using Machine Learning in Sequential Decision
  Making: the Interplay between User Dynamics and Fairness}.
\newblock In {\em Advances in Neural Information Processing Systems}, pages
  15243--15252, 2019.

\bibitem{mouzannar}
{Mouzannar, Hussein and Ohannessian, Mesrob I. and Srebro, Nathan}.
\newblock From fair decision making to social equality.
\newblock In {\em Proceedings of the Conference on Fairness, Accountability,
  and Transparency}, FAT* ’19, page 359–368, New York, NY, USA, 2019.
  Association for Computing Machinery.

\bibitem{disparat-equi}
Lydia~T. Liu, Ashia Wilson, Nika Haghtalab, Adam~Tauman Kalai, Christian Borgs,
  and Jennifer Chayes.
\newblock The disparate equilibria of algorithmic decision making when
  individuals invest rationally.
\newblock In {\em Proceedings of the 2020 Conference on Fairness,
  Accountability, and Transparency}, FAT* ’20, page 381–391, New York, NY,
  USA, 2020. Association for Computing Machinery.

\bibitem{emelianov2019price}
Vitalii Emelianov, George Arvanitakis, Nicolas Gast, Krishna Gummadi, and
  Patrick Loiseau.
\newblock {The Price of Local Fairness in Multistage Selection}.
\newblock {\em arXiv preprint arXiv:1906.06613}, 2019.

\bibitem{fairness-static}
Alexander D’Amour, Hansa Srinivasan, James Atwood, Pallavi Baljekar,
  D.~Sculley, and Yoni Halpern.
\newblock Fairness is not static: Deeper understanding of long term fairness
  via simulation studies.
\newblock In {\em Proceedings of the 2020 Conference on Fairness,
  Accountability, and Transparency}, FAT* ’20, page 525–534, New York, NY,
  USA, 2020. Association for Computing Machinery.

\bibitem{jabbari2017fairness}
Shahin Jabbari, Matthew Joseph, Michael Kearns, Jamie Morgenstern, and Aaron
  Roth.
\newblock {Fairness in reinforcement learning}.
\newblock In {\em Proceedings of the 34th International Conference on Machine
  Learning-Volume 70}, pages 1617--1626, 2017.

\bibitem{NIPS2016_6355}
Matthew Joseph, Michael Kearns, Jamie~H Morgenstern, and Aaron Roth.
\newblock {Fairness in Learning: Classic and Contextual Bandits}.
\newblock In D.~D. Lee, M.~Sugiyama, U.~V. Luxburg, I.~Guyon, and R.~Garnett,
  editors, {\em Advances in Neural Information Processing Systems 29}, pages
  325--333. Curran Associates, Inc., 2016.

\bibitem{wen2019fairness}
Min Wen, Osbert Bastani, and Ufuk Topcu.
\newblock {Fairness with Dynamics}.
\newblock {\em arXiv preprint arXiv:1901.08568}, 2019.

\bibitem{stokey1989recursive}
Nancy~L Stokey.
\newblock {\em {Recursive methods in economic dynamics}}.
\newblock Harvard University Press, 1989.

\bibitem{ljungqvist2018recursive}
Lars Ljungqvist and Thomas~J Sargent.
\newblock {\em {Recursive macroeconomic theory}}.
\newblock MIT press, 2018.

\bibitem{benhabib2015wealth}
Jess Benhabib, Alberto Bisin, and Shenghao Zhu.
\newblock {The wealth distribution in Bewley economies with capital income
  risk}.
\newblock {\em Journal of Economic Theory}, 159:489--515, 2015.

\bibitem{stachurski2019impossibility}
John Stachurski and Alexis~Akira Toda.
\newblock {An impossibility theorem for wealth in heterogeneous-agent models
  with limited heterogeneity}.
\newblock {\em Journal of Economic Theory}, 182:1--24, 2019.

\bibitem{ma2020income}
Qingyin Ma, John Stachurski, and Alexis~Akira Toda.
\newblock {The income fluctuation problem and the evolution of wealth}.
\newblock {\em Journal of Economic Theory}, 187:105003, 2020.

\bibitem{johnson2020precarious}
Nathan~R Johnson and Meredith~A Johnson.
\newblock {Precarious Data: Affect, Infrastructure, and Public Education}.
\newblock {\em Rhetoric Society Quarterly}, 50(5):368--382, 2020.

\bibitem{puar2012precarity}
Jasbir Puar.
\newblock {Precarity Talk: A Virtual Roundtable with Lauren Berlant, Judith
  Butler, Bojana Cveji{\'c}, Isabell Lorey, Jasbir Puar, and Ana
  Vujanovi{\'c}}.
\newblock {\em TDR/The Drama Review}, 56(4):163--177, 2012.

\bibitem{pelletier2020measuring}
David Pelletier, Simona Bignami-Van~Assche, and Ana{\"\i}s Simard-Gendron.
\newblock {Measuring life course complexity with dynamic sequence analysis}.
\newblock {\em Social Indicators Research}, 152(3):1127--1151, 2020.

\bibitem{zheng2020ai}
Stephan Zheng, Alexander Trott, Sunil Srinivasa, Nikhil Naik, Melvin Gruesbeck,
  David~C Parkes, and Richard Socher.
\newblock {The AI economist: Improving equality and productivity with AI-driven
  tax policies}.
\newblock {\em arXiv preprint arXiv:2004.13332}, 2020.

\bibitem{flood2020integrated}
Sarah Flood, Miriam King, Renae Rodgers, Steven Ruggles, and J~Robert Warren.
\newblock {Integrated public use microdata series, current population survey:
  version 7.0 [dataset]}.
\newblock {\em Minneapolis, MN: IPUMS}, 10:D030, 2020.

\bibitem{DQYDJ}
{DQYDJ}.
\newblock [Online] Available:
  \url{https://dqydj.com/average-median-top-household-income-percentiles/}.

\bibitem{census}
{The Census Bureau CPS Annual Social and Economic (March) Supplement 2019}.
\newblock [Online] Available:
  \url{https://www.census.gov/programs-surveys/cps.html}.

\bibitem{preston1975changing}
Samuel~H Preston.
\newblock {The changing relation between mortality and level of economic
  development}.
\newblock {\em Population studies}, 29(2):231--248, 1975.

\bibitem{doi:10.1146/annurev.publhealth.21.1.543}
Adam Wagstaff and Eddy van Doorslaer.
\newblock Income inequality and health: What does the literature tell us?
\newblock {\em Annual Review of Public Health}, 21(1):543--567, 2000.
\newblock PMID: 10884964.

\bibitem{deaton2003health}
Angus Deaton.
\newblock {Health, inequality, and economic development}.
\newblock {\em Journal of Economic Literature}, 41(1):113--158, 2003.

\bibitem{sargent2014quantitative}
Thomas Sargent and John Stachurski.
\newblock {Quantitative economics}.
\newblock Technical report, Citeseer,
  \url{http://citeseerx.ist.psu.edu/viewdoc/download?doi=10.1.1.384.8014&rep=rep1&type=pdf},
  2014.

\bibitem{deaton1989saving}
Angus Deaton.
\newblock {Saving and liquidity constraints}.
\newblock Technical report, National Bureau of Economic Research, 1989.

\bibitem{den2010comparison}
Wouter~J Den~Haan.
\newblock {Comparison of solutions to the incomplete markets model with
  aggregate uncertainty}.
\newblock {\em Journal of Economic Dynamics and Control}, 34(1):4--27, 2010.

\bibitem{kuhn2013recursive}
Moritz Kuhn.
\newblock {Recursive Equilibria in an Aiyagari-Style Economy with Permanent
  Income Shocks}.
\newblock {\em International Economic Review}, 54(3):807--835, 2013.

\bibitem{rabault2002borrowing}
Guillaume Rabault.
\newblock {When do borrowing constraints bind? Some new results on the income
  fluctuation problem}.
\newblock {\em Journal of Economic Dynamics and Control}, 26(2):217--245, 2002.

\bibitem{reiter2009solving}
Michael Reiter.
\newblock {Solving heterogeneous-agent models by projection and perturbation}.
\newblock {\em Journal of Economic Dynamics and Control}, 33(3):649--665, 2009.

\bibitem{schechtman1977some}
Jack Schechtman and Vera~LS Escudero.
\newblock {Some results on “An income fluctuation problem”}.
\newblock {\em Journal of Economic Theory}, 16(2):151--166, 1977.

\bibitem{wakker2008explaining}
Peter~P. Wakker.
\newblock {Explaining the characteristics of the power (CRRA) utility family}.
\newblock {\em Health Economics}, 17(12):1329--1344, 2008.

\bibitem{o2018modeling}
Ted O'Donoghue and Jason Somerville.
\newblock {Modeling risk aversion in economics}.
\newblock {\em Journal of Economic Perspectives}, 32(2):91--114, 2018.

\bibitem{income-precarity}
Michael Nau and Matthew Soener.
\newblock {Income precarity and the financial crisis}.
\newblock {\em Socio-Economic Review}, 17(3):523--544, 06 2017.

\bibitem{baradaran2019color}
Mehrsa Baradaran.
\newblock {\em {The Color of Money: Black Banks and the Racial Wealth Gap}}.
\newblock Belknap Press of Harvard University Press, 2019.

\bibitem{abbasi2019fairness}
Mohsen Abbasi, Sorelle~A Friedler, Carlos Scheidegger, and Suresh
  Venkatasubramanian.
\newblock {Fairness in representation: quantifying stereotyping as a
  representational harm}.
\newblock In {\em Proceedings of the 2019 SIAM International Conference on Data
  Mining}, pages 801--809. SIAM, 2019.

\bibitem{gupta2019equalizing}
Vivek Gupta, Pegah Nokhiz, Chitradeep~Dutta Roy, and Suresh Venkatasubramanian.
\newblock {Equalizing recourse across groups}.
\newblock {\em arXiv preprint arXiv:1909.03166}, 2019.

\bibitem{blau2017gender}
Francine~D. Blau and Lawrence~M. Kahn.
\newblock {The gender wage gap: Extent, trends, and explanations}.
\newblock {\em Journal of Economic Literature}, 55(3):789--865, 2017.

\bibitem{whittaker2019disability}
Meredith Whittaker, Meryl Alper, Cynthia~L Bennett, Sara Hendren, Liz Kaziunas,
  Mara Mills, Meredith~Ringel Morris, Joy Rankin, Emily Rogers, Marcel Salas,
  and Sarah~Myers West.
\newblock {Disability, Bias, and AI}.
\newblock {\em AI Now Institute}, Nov. 2019.

\bibitem{zwerling1992race}
Craig Zwerling and Hilary Silver.
\newblock {Race and job dismissals in a federal bureaucracy}.
\newblock {\em American Sociological Review}, pages 651--660, 1992.

\bibitem{hicken2014racial}
Margaret~T. Hicken, Hedwig Lee, Jeffrey Morenoff, James~S. House, and David~R.
  Williams.
\newblock {Racial/ethnic disparities in hypertension prevalence: reconsidering
  the role of chronic stress}.
\newblock {\em American Journal of Public Health}, 104(1):117--123, 2014.

\end{thebibliography}

% \end{document}\\
\newpage

\appendix
\section{MDP Details}
\label{sec:supp-mcmc}

In this section, we describe the exact values we choose for the transitions between the MDP states. The transition matrices follow a financial security logic. That is, the more income a household has, the more financially secure they will be, meaning even if a higher income household incurs a financial shock, they have the means (financial and non-financial assets) to cover for their loss. Lower and to a less noticeable degree the middle income households, on the other hand, are less secure, and given a positive decision, they have more chances of staying in the same state due to the previous liabilities. And with a negative decision, they tend to move to inferior states with higher probabilities compared to higher-income households. We set the middle-income classes' chances to a random 50\% chance in most cases to give them better chances of improvement. A household can locally reason to go to worse states, better states, or stay in the same states (due to previous debts) as decisions are assigned to them.

The three inferior states given a financial shock due to a negative outcome are: 

\begin{enumerate}
    \item Burning out their savings (encompassed in their net worth),
    \item  Selling one of their non-financial assets which are health-related (e.g., selling their vehicle or house and falling back on public transportation in the midst of the pandemic, which results in a reduction in their health index). Note that since the net worth encompasses both financial and non-financial assets, the value of their net worth does not change here. Rather, transiting from a health-related non-financial asset to a financial one will result in a health index reduction by a factor one score (on the scale of the health indexes in the Census Bureau CPS Annual Social and Economic (March) Supplement 2019).
    
    \item Some of the family member opts out of their health insurance to cover for the debts and liabilities, which in turn reduces their health value by a factor of two scores.
\end{enumerate}

\vspace{0.3cm}
The three positive states given a positive decision are:

\begin{enumerate}
    \item Adding the additional income to their net worth
    \item  Opting for a better health care plan which improves their health index by a factor of one score.
    \item Simply consuming more
\end{enumerate} 

The exact values of transition probabilities are as follows: Given a negative transition, the chances of lower-income classes going to a worse state is 29.6\% (we have three options for bad states given a negative outcome) while the chances of them staying in the same state and not incurring any financial shocks is 11.11\%. The middle-income class is given 50\% chance to stay in the same position given a negative (positive) outcome and a 16\% chance of going to either of the inferior (better) states. Higher-income classes have 55\% chances of staying in the same state given a negative outcome and 15\% chances of going to an inferior position for each of the 3 inferior states (similar probabilities for a positive outcome for lower income classes). Given a positive outcome, the higher income class has 11.11\% chances of staying in the same state and 29.6\% chances of going to each of the better states. After each step of the decision making one of these choices is randomly sampled for the household. The income, net worth, and health indexes get updated accordingly, and the agents continue to interact with the system and environment in an alternating loop.

\section{Finding Optimal Consumption in IFP}

\label{sec:supp-optimal-IFP}

The IFP model uses an endogenous grid method (EGM) to find the optimal consumption path. That is, the EGM  necessitates a grid of savings $s_i$ where each saving is the amount of assets with the consumption subtracted. The grid is utilized to interpolate the optimal consumption function. The basis of the grid is on savings because if the assets are not sufficient, the household would consume them all. Else, the savings will be positive (note that the solution which is considered is the origin-based $ a_0 = c_0 = 0 $). Also, if $ s > 0 $, then $ c < a $. This implies that we can forgo the maximum in \ref{5} and solve the following at each $ s_i $:

$$
c_i =
(u')^{-1}
\left\{
    \beta \, \mathbb E_z
    \hat R
    (u' \circ \sigma) \, [\hat R s_i + \hat Y, \, \hat Z]
\right\} 
$$ \label{eq:iter-cons}

The endogenous asset grid can be computed using $ a_i = c_i + s_i $ once we have tuples $ \{s_i, c_i\} $. We can get an approximation of the policy $ (a, z) \mapsto \sigma(a, z) $ by interpolating $ \{a_i, c_i\} $ at each $ z $ (note that $ z \in \mathsf Z $ so it can be paired with $ a_i $).

\vspace{0.3cm}

\paragraph{Model and Implementation Details:} In the current model, the exogeneous state process $ \{Z_t\} $ is a multi-state process and transition matrix $ P $. We will also assume that $ R_t = \exp(a_r \zeta_t + b_r) $ where $ a_r, b_r $ are constants and $ \{ \zeta_t\} $ is i.i.d.\ standard normal. The labor income itself is defined on the state, percentiles of income. 

Using the endogenous grid, and iterating over the interpolations of the optimal consumption functions, until the consumption function converges to a sufficient level, the implementation derives an approximate optimal consumption function. Note that the optimal consumption function gives the consumption value for a given pair of assets and the state. To derive the exact sequence of the consumption from this, we require information on the sequence of states for the specific instance, which we acquire through the defined process following a Markov-like process. The current implementation then uses that information and the approximate optimal consumption function, as well as the basic needs bounds to showcase the consumption path, i.e., sequence of $c_t$s and $a_t$s.

\end{document}